\documentclass[pdflatex,sn-mathphys-num]{sn-jnl}

\usepackage{graphicx}%
\usepackage{multirow}%
\usepackage{amsmath,amssymb,amsfonts}%
\usepackage{amsthm}%
\usepackage{mathrsfs}%
\usepackage[title]{appendix}%
\usepackage{xcolor}%
\usepackage{textcomp}%
\usepackage{manyfoot}%
\usepackage{booktabs}%
\usepackage[ruled, vlined, noend]{algorithm2e} 
\usepackage{amsmath} 
\usepackage{algpseudocode}%
\usepackage{listings}%
\usepackage[mathlines]{lineno}
\usepackage{amsmath} 

\theoremstyle{thmstyleone}%
\theoremstyle{thmstyletwo}%

\theoremstyle{thmstylethree}%

\begin{document}

\title[Article Title]{USF-Net: A Unified Spatiotemporal Fusion Network for Ground-Based Remote Sensing Cloud Image Sequence Extrapolation}


\author[1]{\fnm{First} \sur{Penghui Niu}}\email{qingxinqazxsw@163.com}

\author[2]{\fnm{Second} \sur{Taotao Cai}}\email{taotao.cai@usq.edu.au}

\author*[3]{\fnm{Third} \sur{Suqi Zhang}}\email{suqizhang@tjcu.edu.cn}

\author*[1,4]{\fnm{Fourth} \sur{Junhua Gu}}\email{jhguhebut@163.com}

\author[1,4]{\fnm{Fifth} \sur{Ping Zhang}}\email{zhangping@hebut.edu.cn}

\author[5]{\fnm{Sixth} \sur{Qiqi Liu}}\email{qiqi6770304@gmail.com}

\author[6]{\fnm{Seventh} \sur{Jianxin Li}}\email{jianxin.li@ecu.edu.au}

\affil*[1]{\orgdiv{School of Artificial Intelligence}, \orgname{Hebei University of Technology}, \orgaddress{\street{} \city{Tianjin}, \postcode{300401}, \state{}\country{China}}}

\affil[2]{\orgdiv{University of Southern Queensland}, \orgname{Organization},\orgaddress{\street{} \city{Toowoomba}, \postcode{487-535}, \state{}\country{Australia}}}

\affil*[3]{\orgdiv{School of Information Engineering}, \orgname{Tianjin University of Commerce}, \orgaddress{\street{}\city{Tianjin}, \postcode{300134}, \state{}\country{China}}}

\affil*[4]{\orgdiv{Hebei Province Key Laboratory of Big Data Calculation}, \orgname{Hebei University of Technology}, \orgaddress{\street{}\city{Tianjin}, \postcode{300401}, \state{}\country{China}}}

\affil[5]{\orgdiv{Trustworthy and General AI Lab, School of Engineering}, \orgname{Westlake University}, \orgaddress{\street{}\city{Hangzhou}, \postcode{310030}, \state{}\country{China}}}

\affil[6]{\orgdiv{Discipline of Business Systems and Operations, School of Business and Law}, \orgname{Edith Cowan University}, \orgaddress{\street{}\city{Joondalup}, \postcode{WA 6027}, \state{}\country{Australia}}}


\abstract{Accurate extrapolation of ground-based cloud imagery is foundational to the stability of photovoltaic (PV) power generation systems. However, current methods rely on static kernels that neglect multi-scale cloud dynamics and lack robust temporal guidance for long-range dependency modeling. Moreover, the quadratic computational cost of standard attention mechanisms impedes real-time deployment. To bridge these gaps, we propose the Unified Spatiotemporal Fusion Network (USF-Net), which integrates adaptive large-kernel convolutions with a low-complexity attention mechanism. Specifically, USF-Net incorporates a unified spatiotemporal module (USTM), which comprises a spatial information branch equipped with a spatial selection module (SSM) for dynamic multi-scale context extraction and a temporal information branch featuring a temporal agent attention module (TAM). The TAM efficiently models long-range temporal dependencies, circumventing the computational bottlenecks of traditional attention. In the decoder, a dynamic update module (DUM) leverages initial temporal states to preserve motion signatures, thereby mitigating ghosting effects. Crucially, we introduce the \textit{ASI-Cloud Image Sequence (ASI-CIS) dataset}, a large-scale, high-resolution benchmark designed to address current data limitations. Extensive experiments on ASI-CIS demonstrate that USF-Net establishes a new state-of-the-art, offering a superior trade-off between prediction accuracy and computational efficiency for ground-based cloud forecasting. The dataset and source code will be available at https://github.com/she1110/ASI-CIS.}




\maketitle

\section{Introduction}\label{sec:intro}
Driven by the imperative to decarbonize the global energy landscape, photovoltaic (PV) power generation has established itself as a cornerstone of renewable energy transitions, distinguished by its zero-carbon emissions and ubiquitous resource availability~\cite{Intro1}. However, the integration of PV systems into the electrical grid is complicated by the stochastic nature of solar irradiance. Rapid fluctuations in power output, frequently precipitated by highly dynamic cloud cover, impose severe stress on power dispatch systems and necessitate robust energy storage configurations to maintain grid stability~\cite{Intro2}. Consequently, high-resolution solar irradiance forecasting has become a prerequisite for enhancing PV integration capacity and ensuring power system reliability.

Since PV output exhibits a strong correlation with solar irradiance, which is primarily modulated by atmospheric cloud dynamics, precise cloud observation is critical~\cite{Intro3}. Irradiance variability is generally categorized into sustained shading (e.g., extensive stratiform coverage) and transient shading (e.g., rapid cumulonimbus convection). While numerical weather prediction (NWP) models effectively forecast sustained events, they lack the temporal granularity required for transient shading~\cite{Intro4}. Conversely, satellite imagery, though offering broad coverage, typically suffers from coarse spatial resolution ($> $1 km) and high latency ($\ge $30 minutes), rendering it inadequate for tracking localized, micro-scale cloud evolution~\cite{Intro5, Intro6}. In contrast, ground-based sky imagers capture cloud structure and motion with high spatiotemporal resolution, making them the optimal modality for ultra-short-term irradiance forecasting~\cite{25TIM}. Therefore, the accurate extrapolation of ground-based cloud image sequences, essentially a deterministic spatiotemporal prediction problem, is pivotal for operational grid stability~\cite{Intro7, Intro8, Intro9}.

Advancing the state-of-the-art (SOTA) in cloud extrapolation requires addressing two fundamental challenges: (a) modeling the complex, nonlinear, and multi-scale morphological deformations of clouds, and (b) efficiently capturing long-range spatiotemporal dependencies to satisfy the real-time inference constraints of industrial applications.

Existing methodologies fall into two primary paradigms: traditional optical flow frameworks and deep learning-based approaches. Traditional methods, often relying on vector field representations and similarity maximization, are frequently hampered by excessive computational overhead and limited capacity to represent nonlinear features, leading to performance degradation under complex atmospheric conditions~\cite{Intro10, Intro11}.

The advent of deep learning has shifted the paradigm toward data-driven spatiotemporal modeling. Recurrent Neural Networks (RNNs) ~\cite{Intro12} and their variants, such as Long Short-Term Memory (LSTM) networks~\cite{Intro14}, have been widely adopted for their ability to model temporal dynamics. To address the spatial limitations of standard RNNs, hybrid architectures like the Convolutional LSTM (ConvLSTM) were introduced to integrate spatial feature extraction with temporal memory~\cite{ConvLSTM}. Subsequent innovations have focused on multi-scale convolutional kernels to capture hierarchical cloud morphology and attention mechanisms to model global dependencies~\cite{Intro16, Intro17}.

Despite these advancements, significant limitations persist. First, the reliance on static, fixed-size kernels prevent models from dynamically adapting their receptive fields (RF) to the diverse scales of cloud structures. Second, the interaction between spatial and temporal feature streams is often decoupled, lacking an explicit mechanism for temporal flow to guide spatial feature learning. Third, the quadratic complexity of standard self-attention mechanisms often creates a computational bottleneck prohibitive for high-resolution, real-time monitoring. Finally, the ghosting effect, a blurring artifact prevalent in prediction decoders, remains a critical issue, often exacerbated by the decay of contextual information during upsampling.

Furthermore, current research is constrained by the quality of available benchmarks. Existing datasets often suffer from low spatial resolution and hardware-induced visual artifacts, which introduce noise and limit the applicability of extrapolation models to real-world PV forecasting~\cite{MSTANet, CloudPredRNN++}. Thus, there is a pressing need for a high-resolution, multi-scale, and cross-seasonal benchmark dataset.

To bridge these gaps, this article proposes the Unified Spatiotemporal Fusion Network (USF-Net), a novel encoder-decoder architecture that integrates adaptive large-kernel convolutions with a low-complexity attention mechanism. Specifically, the encoder employs depthwise separable (DW) convolutions and squeeze-excitation (SE) blocks for efficient multi-scale feature extraction. Following the encoder, we propose a unified spatiotemporal module (USTM) comprising three core components: 1) a spatial information branch, in which a spatial selection module (SSM) is designed for dynamic receptive field adjustment; 2) a temporal information branch incorporating a temporal agent attention module (TAM) is introduced for efficient dependency modeling; and 3) a dynamic spatiotemporal module: a temporal guidance module (TGM) is proposed to explicitly fuse temporal flow with spatial features. For the decoder part, a dynamic update module (DUM) is implemented to mitigate the ghosting effect by using the initial temporal state as a gating operator to reweight spatiotemporal features while preserving high-frequency motion signatures. Additionally, we introduce and publicly release the ASI-CIS dataset, a large-scale, high-resolution benchmark for ground-based cloud extrapolation. The main contributions of this work are summarized as follows:
\begin{enumerate}
	\item{A novel unified spatiotemporal architecture, USF-Net, is proposed. It explicitly integrates temporal flow information to guide spatial feature learning, which enhances the coherence of temporal-spatial dependencies modeling and significantly improves prediction accuracy for complex cloud dynamics.}
	\item{A USTM is designed to serve as the core of the network. It features a SSM for dynamic, adaptive multi-scale feature extraction and a low-complexity TAM that effectively balances predictive accuracy with computational efficiency.}
	\item{A DUM is introduced in the decoder. It leverages initial temporal states as an attention operator to suppress the ghosting effect and preserve fine-grained cloud details.}
	\item{The introduction and public release of the ASI-CIS dataset. As a major contribution to the community, ASI-CIS is a newly introduced, large-scale, high-resolution, multi-seasonal benchmark that addresses key limitations of previous datasets. It offers a more realistic and challenging foundation for advancing ground-based cloud extrapolation models. Extensive experiments on ASI-CIS show that USF-Net outperforms state-of-the-art methods in both prediction accuracy and computational efficiency.}
\end{enumerate}
The remainder of this paper is organized as follows. In Section~\ref{sec:related}, we introduce the related works of ground-based cloud image sequence precidtion. In Section~\ref{sec:method}, we introduce the detailed structure of the proposed method. In Section~\ref{sec:exp}, we evaluate the performance of our proposed methods. Finally, Section~\ref{sec:conclusion} concludes the paper.
\section{Related works} \label{sec:related}
Precise cloud image sequence extrapolation underpins the operational stability of grid-connected PV systems. Existing methodologies categorize primarily into two categories: traditional methods and deep learning-based methods.
\subsection{Traditional methods for cloud image sequence extrapolation}
Traditional frameworks predominantly leverage optical flow (OF) algorithms. These techniques estimate the instantaneous velocity field of pixel-wise motion in ground-based cloud images, capturing spatiotemporal trends of cloud dynamics. Several studies have adopted OF-based methods for ground-based remote sensing cloud image sequence extrapolation. Omnidirectional tracking frameworks quantify relationships between cloud motion directionality and velocity magnitudes in cloud dynamics~\cite {Re1, Re2}. Boundary information within cloud imagery constitutes a critical determinant of prediction accuracy. Chang et al. implemented the Horn-Schunck OF algorithm to compute velocity variations for each pixel, augmenting motion field estimation through supplementary information integration~\cite{Re3}. Nevertheless, such methods impose significant computational burdens. Conversely, Wang et al. introduced a mathematical analysis framework to characterize inter-frame disparities, achieving reduced computational resource consumption while maintaining predictive performance~\cite{Re4}.

Despite the operational feasibility of the aforementioned methodologies in executing cloud image sequence extrapolation tasks, conventional approaches exhibit persistent limitations, including being computationally prohibitive and exhibiting limited motion modeling capabilities. These deficiencies manifest in their inability to capture temporal motion patterns under complex cloud regimes characterized effectively. The rapid advancement of DL techniques in computer vision has spurred transformative progress in this domain. Numerous studies have applied DL to the prediction of cloud image sequence extrapolation, achieving remarkable advancements.
\subsection{DL-based methods for cloud image sequence extrapolation}
Historically, recurrent neural networks and their variant long short-term memory networks constituted the foundational approach for temporal sequence prediction. Several studies have successfully employed LSTM architecture to capture long-range dependencies in spatiotemporal sequences. However, standalone LSTM models exhibit elevated computational costs while struggling to extract complex spatial information effectively. Consequently, numerous works integrate CNN with LSTM frameworks to jointly extract localized spatial features and temporal information for sequential motion prediction. Subsequent architectures augmented the standard Convolutional LSTM by introducing a bidirectional memory propagation mechanism to model short-term spatiotemporal dynamics~\cite{PredRNN}. Subsequently, Further iterations incorporated gradient highway units to alleviate vanishing gradients in LSTM-based models while integrating Causal-LSTM modules to strengthen spatial feature representation and short-term temporal modeling~\cite{PredRNN++}. Li et al. proposed cascaded Causal-LSTM layers to improve short-term prediction accuracy for cloud imagery. The model was augmented by GHUs with auxiliary skip connections to enhance spatiotemporal uniformity in modeling~\cite{CloudPredRNN++}. However, ground-based remote sensing cloud imagery characterizes by high-resolution cloud formations with multi-scale variations under complex meteorological conditions. Existing methodologies remain suboptimal for cloud sequence extrapolation tasks. To resolve multi-scale dynamic states of cloud clusters at varying scales within cloud imagery, several studies have integrated multi-scale convolutional kernels. For instance, MSSTNet employs 3D convolutions with diverse kernel sizes to enhance the capacity of the model for multi-resolution image forecasting~\cite{Re5}. Wang et al. introduced 3D tensor augmentations within LSTM architectures to expand the effective receptive field~\cite{Re6}. However, the adoption of 3D convolution operations incurs significant computational overhead. The MSTANet employs multi-scale large kernels to aggregate multi-scale contextual information from cloud imagery while leveraging depthwise separable convolutions to construct large kernels with reduced computational complexity~\cite{MSTANet}. Accurate cloud image sequence extrapolation serves as a critical factor in ultra-short-term PV power forecasting, and the modeling of long-range spatiotemporal dependencies becomes particularly paramount. To this end, Chang et al. designed an MAU that simultaneously enlarges the model’s receptive field and captures spatial motion patterns across cloud sequences~\cite{MAU}. The Motion RNN introduces a Motion RGU module to unify transient variation modeling and motion trend representation~\cite{Re7}. When embedded within RNN architectures, this approach significantly improves spatiotemporal prediction accuracy under complex meteorological scenarios.

Attention mechanisms have proven effective for establishing long-range dependencies, facilitating the extraction of temporal features in sequence prediction tasks. The STANet introduces a context gating unit (CGU) as an attention mechanism to unify the modeling of instantaneous cloud characteristics and motion trends~\cite{STANet}. Similarly, SimvPV2 incorporates a gated spatiotemporal attention module to enforce spatiotemporal consistency in sequence modeling~\cite{Re8}. Tan et al. decomposed temporal attention into intra-frame static attention and inter-frame dynamic attention through a dedicated temporal attention unit, capturing spatial features and temporal correlations~\cite{TAU}. Li et al. integrated a self-attention memory unit into the Cascaded Causal LSTM (CCLSTM) framework to extract long-term dependencies, enabling spatiotemporal modeling for cloud sequence prediction~\cite{CCLSTM}. Furthermore, the MSTANet introduces a multi-scale temporal attention mechanism that combines local temporal variations and global temporal variations, significantly enhancing the network's temporal modeling capacity for cloud image extrapolation tasks~\cite{MSTANet}.

While demonstrating efficacy, these methodologies encounter three distinct limitations. First, these approaches solely employ multi-scale convolutional kernels to capture contextual information, lacking adaptive mechanisms to extract features at varying resolutions dynamically. In addition, during spatiotemporal dependency modeling, the absence of temporal guidance hinders the unified integration of spatial and temporal information, resulting in suboptimal long-term dependency capture. Furthermore, existing attention mechanisms for temporal feature extraction neglect to balance computational complexity with prediction accuracy, leading to inefficiencies in practical deployment.
\begin{figure}[!t]
	\centering{\includegraphics[width=3.5in]{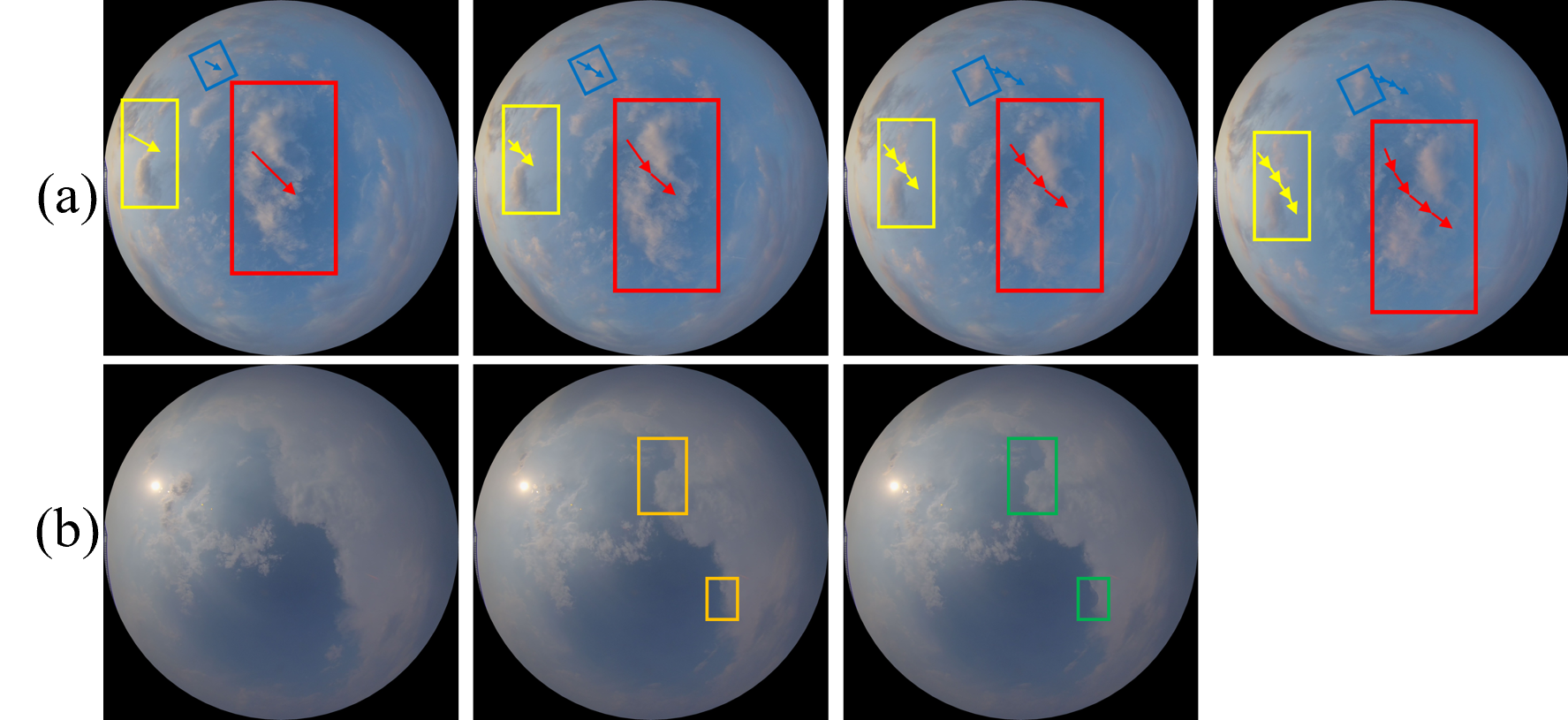}}
	\caption{(a) illustrates multi-scale cloud movement. The red, yellow, and blue blocks represent displacement vectors of large, medium, and small-scale clouds, respectively. The arrow indicates the direction of the movement trend. (b) demonstrates “ghosting effects” in cloud image sequence extrapolation. The orange block denotes ground truth (GT), whereas the green block indicates extrapolation results.}
	\label{fig_0}
\end{figure}
\begin{algorithm}[t!]
	\setlength{\intextsep}{0pt}
	\setlength{\abovedisplayskip}{0pt}
	\setlength{\belowdisplayskip}{0pt}
	\caption{Procedure of the USF-Net}
	\label{alg:usfnet}
	\SetKwInOut{Input}{Input}
	\SetKwInOut{Output}{Output}
	\SetKwRepeat{Repeat}{repeat}{until}
	
	\Input{Input cloud sequence $X_{t}^{T}=\{x_i\}_{t+1}^{T}$}
	\Output{Extrapolated sequence $Y_{T+1}^{T+\tau}=\{y_i\}_{T+1}^{T+\tau}$}
	\BlankLine
	
	\Repeat{convergence}{
		Let layer $i$ = 1, $loss$ = 0.0
		
		$L=\lambda_{1}L_{M}+\lambda_{2}L_{MS}+\lambda_{3}L_{C}$ ($L$: loss function)
		
		\For{$i \leftarrow 1$ \KwTo $3$}{
			$X_B \leftarrow N_i(X_{t}^{T})$ \tcp*{Encoder layer $i$}
			\If{$i = 3$}{ $X_T \leftarrow X_B$ }
		}
		
		$X_S \leftarrow \text{SiB}(X_B)$ 
		
		$X_T \leftarrow \text{TiB}(X_T)$ 
		
		$X_D \leftarrow \text{DSM}(X_S, X_T)$ 
		
		$D_4 \leftarrow \text{DUM}(X_D, X_{T_0})$ 
		
		\For{$k \leftarrow 3$ \KwTo $1$}{
			$D_k \leftarrow \text{UP}(D_{k+1})$ \tcp*{Decoder layer $k$}
		}
		$Y \leftarrow \text{Conv}_{1\times1}(D_1)$ 
		
		$\mathcal{L} \leftarrow \lambda_1 L_M + \lambda_2 L_{MS} + \lambda_3 L_C$ 
		
		$\theta \leftarrow \theta - \nabla_\theta \mathcal{L}$ \tcp*{Parameter update}
	}
	\Return $Y_{T+1}^{T+\tau}$
\end{algorithm}
\vspace*{-10pt}
\section{Proposed Method}\label{sec:method}
This section elucidates the architecture of USF-Net, commencing with the mathematical formulation of the extrapolation task. Subsequently, we detail the novel architectural components: the unified spatiotemporal module, the dynamic update module, and the composite loss function designed to enforce structural fidelity.
\subsection{Formulation and architectural overview}
The ground-based cloud image sequence extrapolation task operates on an input sequence of $T$ historical frames, denoted as $X_{t}^{T}=\left \{ x_{i}  \right \}  _{t+1}^{T}$, where each frame  $x_{i}\in \mathbb{R} ^{C\times H\times W}$. The objective is to predict a subsequent sequence of $\tau$ future frames after $T$, $Y_{T+1}^{T+\tau}=\left \{ y_{i}  \right \}  _{T+1}^{T+\tau}$, where $y_{i}\in \mathbb{R} ^{C\times H\times W}$. Here, $C$, $H$, and $W$ represent the channel, height, and width dimensions, respectively. The model, parameterized by $\theta$, approximates a mapping function $F_{\theta}: X_{t}^{T} \rightarrow Y_{T+1}^{T+\tau}$. This mapping is optimized by maximizing the log-likelihood of the predicted cloud frames relative to their ground-truth counterparts. Formally, the optimization objective is defined as:
\begin{equation}
	\theta _{T}=arg\underset{\theta }{max }  {\textstyle \sum_{i=T+1}^{T+\tau}} logP\left (y_{i} \mid x_{i};\theta  \right ).
	\label{equ_0}
\end{equation}
\par\noindent where $\theta_{T}$ represents the learnable parameters that align the predicted distribution with the physical reality of cloud formation dynamics.

Existing methods have demonstrated that the ground-based cloud image sequence extrapolation task faces several critical challenges. As illustrated in Fig.~\ref{fig_0} (a), the scale-variant properties of cloud formations during motion introduce inaccuracies in sequence extrapolation due to multi-scale variations. Moreover, Fig.~\ref{fig_0} (b) reveals that prevalent approaches suffer from partial contextual information loss during the decoder phase, leading to the emergence of “ghosting effects” that complicate cloud motion trajectory prediction. These motivate the design of a multi-scale network model with a spatiotemporally unified architecture, aiming to improve the precision of cloud sequence prediction while simultaneously enhancing inference efficiency.
\begin{figure*}[!t]
	\centering{\includegraphics[width=5in]{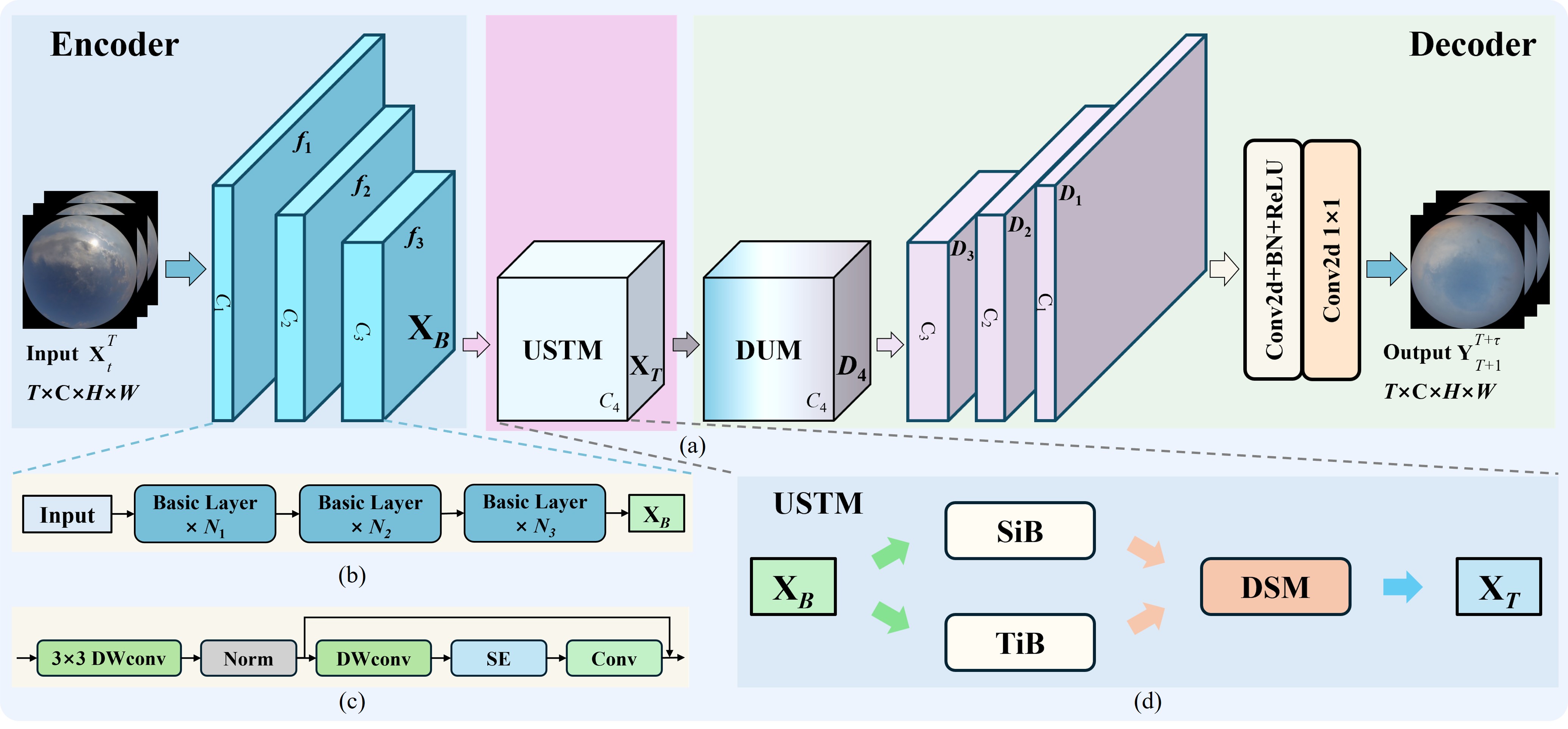}}
	\caption{(a) The structure of the proposed USF-Net is composed of three parts: the encoder comprises three Basic Layers, the USTM and the decoder comprises a dynamic update module (DUM). $C_{i}$ denotes the channel of the feature map. (b) The structure of the encoder, where $N_{1}$, $N_{2}$, and $N_{3}$ are 2, 2, and 3, respectively. The output of the encoder is $X_{B}$. (c) The specific structure of the Basic Layer. (d) The diagram of the proposed Unified SpatioTemporal Module (USTM) comprises three core components: a spatial information branch (SiB), a temporal information branch (TiB), and a dynamic spatiotemporal module (DSM). The output of the USTM is $X_{T}$.}
	\label{fig_1}
\end{figure*}
\subsection{Overview of structure}
The proposed USF-Net adopts a sophisticated encoder-decoder configuration (Fig. \ref{fig_1}a). The procedure of our USF-Net is described in Algorithm \ref{alg:usfnet}. To mitigate the inaccuracies arising from the scale-variant nature of cloud motion and the ghosting effects prevalent in existing decoders , USF-Net integrates a USTM at the bottleneck.

In the encoder, as shown in Fig. \ref{fig_1} (a) and (b), the input tensor $X_{t}^{T}\in \mathbb{R} ^{B\times T\times C\times H\times W}$ traverses three Basic Layers. Each layer incorporates a $3\times 3$ Depth-Wise (DW) convolution for local feature extraction, followed by layer normalization, a secondary DW convolution, and a Squeeze-and-Excitation (SE) block to recalibrate channel dependencies. Residual connections facilitate gradient propagation, enriching the representation capabilities. We denote the feature map of the $i$-th layer as $f_{i} (i \in [1,2,3])$, with dimensions scaling to $T\times 2^{i-1}\cdot 64\cdot (H\times W)/2^{i+1}$. This process culminates in an intermediate feature map $X_{B} \in \mathbb{R}^{T\cdot C_{3}\cdot (H\times W)/16}$.

The USTM processes $X_{B}$ through dual pathways: a spatial branch that elicits multi-scale spatial features and a temporal branch that captures sequential dependencies. A DSM subsequently fuses these streams, leveraging temporal cues to guide spatial feature reconstruction.

In the decoder, we introduce the DUM to resolve the loss of contextual information often observed during upsampling. The DUM utilizes a gating mechanism to reweight initial temporal features, thereby refining the multi-scale spatiotemporal representations and preserving semantic coherence throughout the extrapolation horizon. Consequently, our architecture can be interpreted as a spatiotemporally unified framework optimized for ground-based cloud image extrapolation, balancing computational efficiency with prediction accuracy. The details of our method are as follows.
\subsection{Unified spatioemporal module (USTM)}
Accurate cloud extrapolation necessitates the simultaneous resolution of multi-scale spatial variations and non-stationary temporal dynamics. Conventional large-scale kernels often lack the adaptability required to capture fine-grained cloud textures, while standard self-attention mechanisms incur prohibitive quadratic computational costs. The USTM addresses these limitations by synergizing a spatial information branch (SiB), a temporal information branch (TiB), and a DSM (Fig. \ref{fig_1}d).
\begin{figure*}[!t]
	\centering{\includegraphics[width=5in]{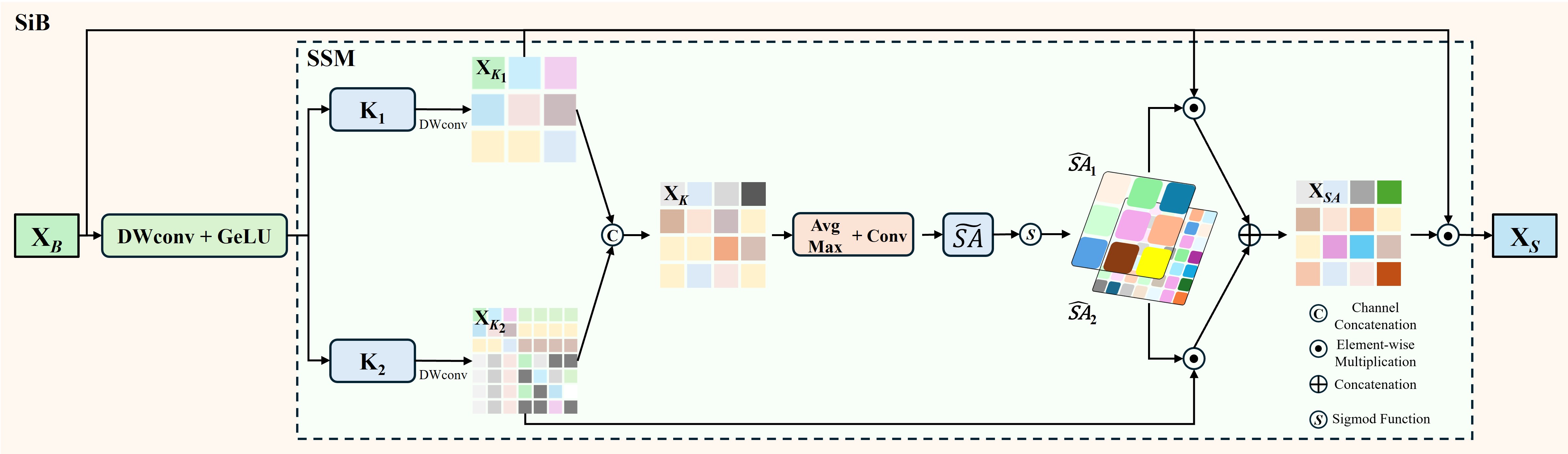}}
	\caption{The structure of the proposed SiB. The SSM employs explicitly decomposed convolution operations to generate varying receptive field sizes, thereby enhancing the network's multi-scale representational capacity.}
	\label{fig_3.0}
\end{figure*}
\subsubsection{Spatial information branch (SiB)}
To accommodate the nonlinear and scale-variant nature of cloud formations, the SiB employs a SSM, as shown in Fig.~\ref{fig_3.0}. Inspired by the Large Kernel Selection (LSK) mechanism~\cite{LSK}, the SSM dynamically adjusts the receptive field to align with the varying scales of cloud structures.

The SSM initially processes the input via a DW convolutional layer and a GeLU activation function, balancing efficiency with nonlinear expressiveness. We employ multi-scale convolutional kernels to adaptively select relevant spatial contexts. Let $K_{i}$ denote the $i$-th kernel; the resulting feature map $X_{K_{i}}$ is derived as:
\begin{equation}
		X_{K_{i}}=\mathrm {DW} \left ( K_{i} \right ).
	\label{equ_1}
\end{equation}
\par\noindent We utilize a decomposition strategy for large kernels, setting $K_{1}=3$ (dilation $d_{1}=1$) and $K_{2}=7$ (dilation $d_{2}=3$). The effective $\mathrm{RF}$ is expanded via:
\begin{flalign}
	\mathrm{RF}_{1}&=k_{1}\\
	\mathrm{RF}_{2}&=d_{2}\left ( k_{2}-1 \right ) +\mathrm{RF}_{1}
	\label{equ_2}
\end{flalign}
\par\noindent
This explicit decomposition mimics a large kernel ($K=21$) while minimizing parameter overhead. The multi-scale features are concatenated to form $X_{K}$, which encapsulates diverse receptive fields. To facilitate cross-spatial feature interaction, we apply channel-wise average and max pooling, followed by a convolution to generate the spatial interaction attention map $\widetilde{SA}$:
\begin{flalign}
	X_{K}&=\mathrm{Cat}\left ( X_{K_{1}}, X_{K_{2}}\right )\\ 
	\widetilde{SA} &=\mathrm{Conv}\left(\mathrm{Avg}\left(X_{K} \right), \mathrm{Max}\left(X_{K}\right)\right).
	\label{equ_3}
\end{flalign}
\par\noindent where $\mathrm {Cat}(\cdot ) $ denotes the concatenation of the channel, $\mathrm {Avg}(\cdot ) $ and $\mathrm {Max}(\cdot ) $ denote the average pooling and max pooling, respectively. 

Attention weights $\widehat{SA}_{i}$ for each scale are computed via a sigmoid function ($\sigma$). These weights modulate the decomposed feature maps, which are subsequently fused and convolved to produce the multi-scale spatial attention map $X_{SA}$. The final output $X_{S}$ is the element-wise product of the input $X_{B}$ and $X_{SA}$:
\begin{flalign}
	\widehat{SA}_{i} &=\sigma \left(\widetilde{SA}_{i}\right) \\
	X_{SA}&=\mathrm{Conv}\left(\mathrm{Concat}\left(\widehat{SA}_{1}\ast X_{K_{1}}, \widehat{SA}_{2}\ast X_{K_{2}}\right)\right) \\
	X_{S} &= X_{SA}\ast X_{B}.
	\label{equ_4}
\end{flalign}
\par\noindent where $\ast$ denotes the element-wise multiplication.
By dynamically adjusting the receptive fields of spatial targets within the spatial branch, the proposed method effectively captures contextual information across varying cloud scales.
\subsubsection{Temporal information branch (TiB)}
Capturing the non-stationary motion of cloud sequences requires robust temporal modeling. Unlike recurrent networks, which suffer from sequential processing bottlenecks, or standard ViTs with quadratic complexity, we propose a TAM. This module integrates the high precision of Softmax attention with the linear complexity of Agent Attention, optimizing the trade-off between computational efficiency and representational capacity.

\begin{figure}[!t]
	\centering
	\includegraphics[width=4in]{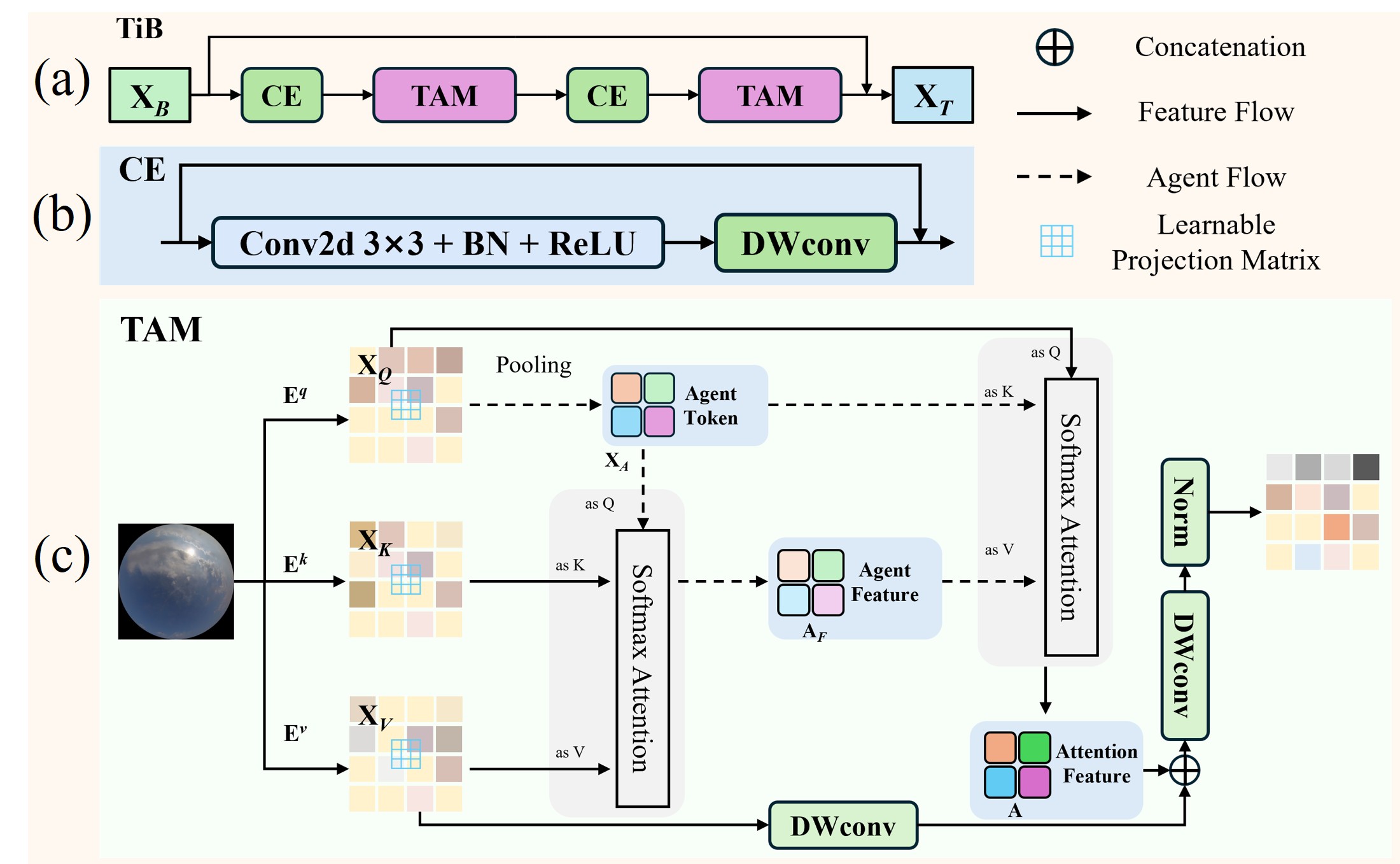}
	\caption{(a) The overall structure of the proposed TiB. (b) The proposed CE consists of a $3\times3$ convolution, batch normalization (BN), ReLU activation, and a DW convolutional layer with residual connections. (c) The proposed DSM, the dashed line denotes the feature flow of Agent attention, and the solid line denotes the feature flow of Softmax attention.}
	\label{fig_3.1}
\end{figure}

As shown in Fig. \ref{fig_3.1}, The TiB comprises a Convolutional Embedding (CE) layer and stacked TAMs. Deviating from standard patch embedding, the CE layer utilizes a CNN to preserve spatial coherence. It consists of a CBR block ($3\times3$ Conv, Batch Normalization, ReLU) followed by a DW convolution with residual connections.

The TAM establishes long-range dependencies without the computational overhead of standard self-attention. Instead of projection matrices, we construct customized convolutional kernels to define adaptive receptive fields for each pixel. The Query ($Q$) and Key ($K$) projections are formulated as:
\begin{flalign}
	X_{Q_{ij}}& =\sum_{l=-1}^{1} \sum_{g=-1}^{1} E_{2+l,2+g}^{q}X_{i+l,j+g} \\
	X_{K_{ij}} &=\sum_{l=-1}^{1} \sum_{g=-1}^{1} E_{2+l,2+g}^{k}X_{i+l,j+g}.
	\label{equ_5}
\end{flalign}
\par\noindent where $E^{q}, E^{k} \in \mathbb{R}^{T\times C\times 3\times 3}$ are learnable matrices aggregating local neighborhoods. To achieve linear complexity, we generate an agent token $X_{A} \in \mathbb{R}^{T\times C\times n}$ ($n \ll N$) via pooling. The attention computation proceeds in two stages: first between the Agent ($Q$) and the global context ($K, V$) to generate agent features $A_{F}$, and second between the global Query ($X_{Q}$) and the Agent ($K$) to distribute the attention:
\begin{flalign}
	X_{A} &=\mathrm{Pooling}\left(X_{Q}\right) \\
	A_{F} &=\mathrm{Soft} \left(X_{A}, \left(X_{K}\right)^{T}\right)X_{V} \\
	A&=\mathrm{Soft} \left(X_{Q}, \left(X_{A}\right)^{T}\right)A_{F}.
\label{equ_6}
\end{flalign}
\par\noindent where $\mathrm{Soft}(\cdot)$ denotes softmax attention function. The resulting feature $A$ undergoes DW convolution and normalization to yield the final temporal embedding $X_{T}$.
\begin{flalign}
	X_{T}&=\mathrm{BN}(\mathrm{DW}(\left(\mathrm{Concat}\left(\mathrm{DW}\left(X_{V}, A\right)\right)\right)))
	\label{equ_7}
\end{flalign}
	\label{equ_7}
\noindent
\begin{figure}[!t]
	\centering
	\includegraphics[width=5in]{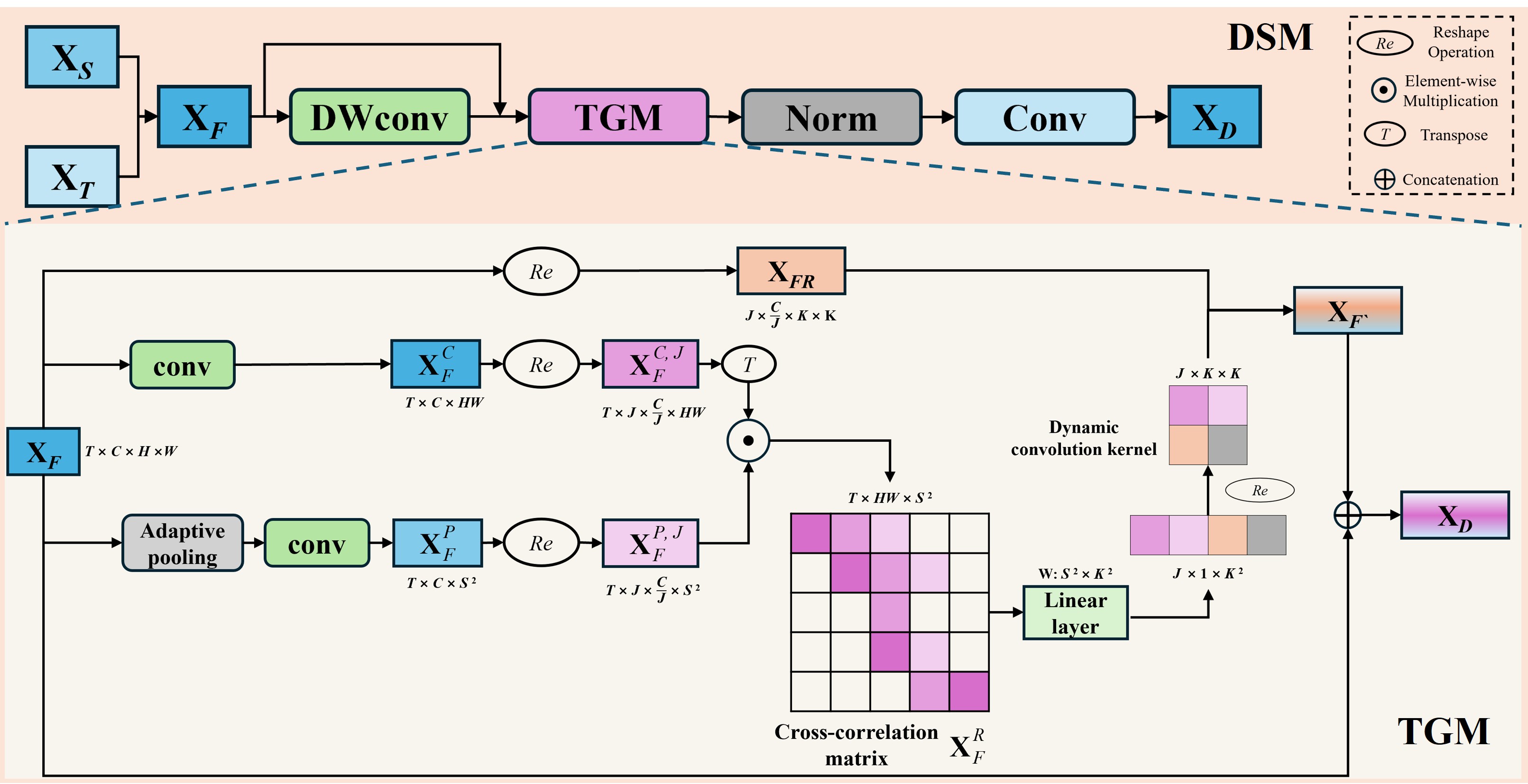}
	\caption{The structure of the proposed DSM. The bottom-hand side of the figure shows the structure of the TGM in detail. The learnable dynamic convolution kernels are generated by applying weighted guidance from temporal flow information to spatial feature maps utilizing temporal flow information.}
	\label{fig_3.2}
\end{figure}
\subsubsection{Dynamic spatiotemporal module (DSM)}
As shown in Fig. \ref{fig_3.2}, the outputs of the spatial and temporal branches are fused to generate a combined feature map $ X_{F}\in\mathbb{R}^{T\times C\times H\times W}$, which is then enhanced via a DW convolutional layer and residual connection for enriched representation. The temporal guidance module is subsequently applied to implement time flow information guidance. Specifically, $ X_{F}$ is split into two components, $ X^{C}_{F}\in\mathbb{R}^{T\times C\times HW}= \mathrm{Conv}\left(X_{F}\right) $ and $ X^{P}_{F}\in\mathbb{R}^{T\times C\times S^{2}}= \mathrm{Conv}\left( Pool\left( X_{F}\right )\right ) $. The $X^{P}_{F}$ undergoes adaptive average pooling to aggregate spatial information into $S$ regions. Then, $ X^{C}_{F}$ and $ X^{C}_{F}$ are divided into $J$ groups along to channels to obtain $ X^{C,J}_{F}\in\mathbb{R}^{T\times J\times \frac{C}{J} \times HW} =Re\left(X_{F}^{C}\right)$ and $ X^{P,J}_{F}\in\mathbb{R}^{T\times J\times \frac{C}{J} \times S^{2}} =Re\left(X_{F}^{C}\right)$, respectively, where $Re\left ( \cdot  \right ) $ denotes a reshape operation. A cross-correlation matrix, $ X^{R}_{F}\in\mathbb{R}^{T\times HW \times S^{2}}$, is computed through matrix multiplication between corresponding groups, capturing inter-region contextual relationships. The key idea is to represent inter-region contextual relationships via $J$ group vectors, enabling the learning of dynamic convolution kernels from $ X^{R}_{F}$. Long-term dependencies are dynamically modulated by propagating contextual information across correlated regions. Subsequently, $J$ dynamic convolution kernels of size $K\times K$ are generated by mapping $ X^{R}_{F}$ through a learnable linear layer $W\in\mathbb{R}^{S^{2}\times K^{2}}$, producing spatiotemporal tokens that encode regional context from the correlation matrix. The feature $ X_{F}$ is also divided into $J$ channel groups, which are then convolved with the reshaped kernels to share spatiotemporal dependencies, yielding the dynamically modulated feature $X_{F'}$. The output $X_{U}$ of the TGM is obtained by combining $ X_{F}$ and $X_{F'}$. Finally, $X_{U}$ is processed through a normalization layer and a convolutional layer to generate the output $X_{D}$ of the dynamic spatiotemporal module.
\begin{figure}[!t]
	\centering
	\includegraphics[width=4in]{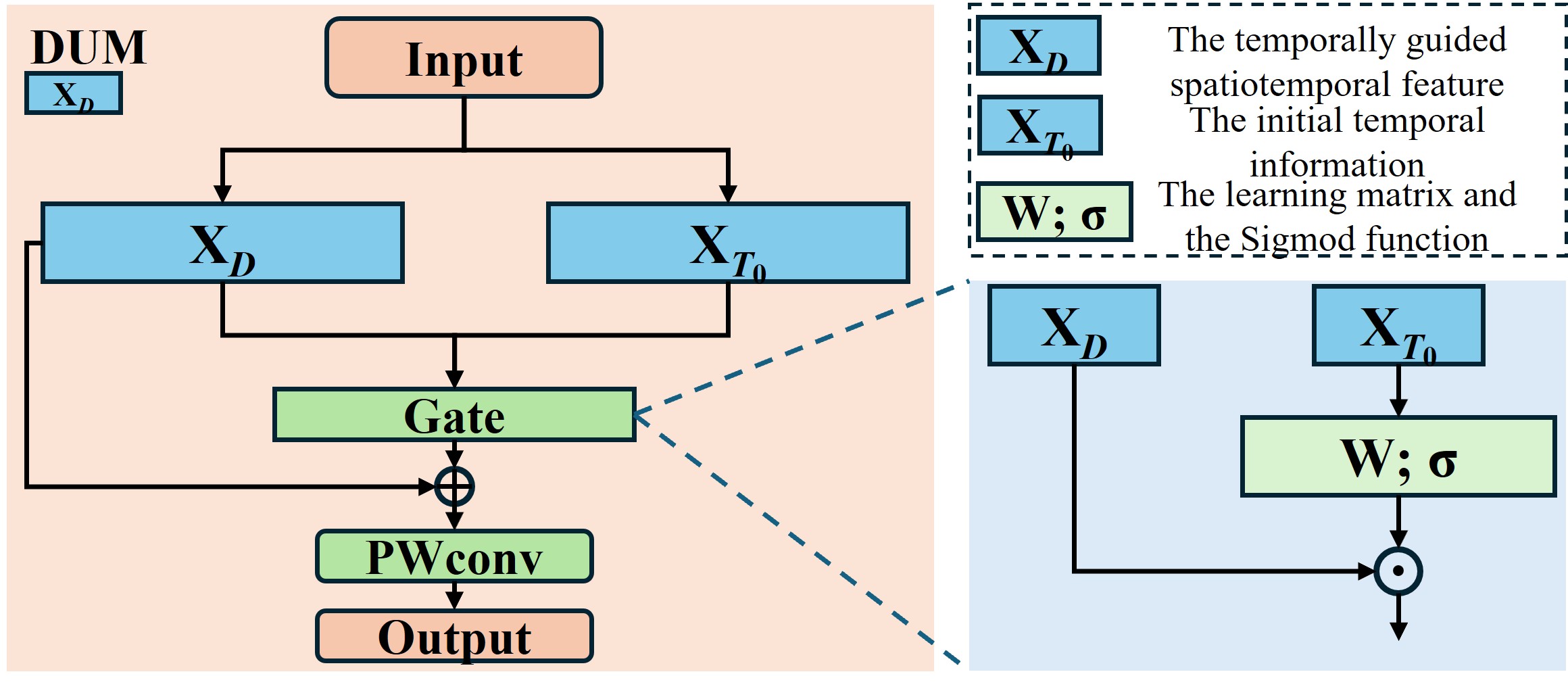}
	\caption{The structure of the proposed DUM. The $X_{e}$ and $X_{M}$ on the left side of the figure are the temporally guided spatiotemporal feature and the initial temporal information from TAM, respectively. The $W$ and $\sigma$ on the right side represent the matrix and active function in the gate unit, respectively.}
	\label{fig_3.3}
\end{figure}
\subsection{Dynamic update decoding}
A pervasive challenge in cloud sequence extrapolation is the ghosting effect, characterized by blurred predictions due to information decay in the decoder. Standard decoders using simple lateral connections often fail to preserve high-frequency temporal details. To rectify this, we introduce a decoder equipped with DUM.

The DUM functions as a gated recurrent unit that reinforces the temporally guided context $X_{D}$ (from USTM) with the initial temporal information $X_{T_{0}}$ from the TAM, as shown in Fig. \ref{fig_3.3}. This mechanism prevents the dilution of temporal cues during upsampling. Then, the features of the two branches are fused by a pointwise (PW) convolution to refine the fused information. The global features $D_{4}$ with temporal flow information are generated through the dynamic update decoding, which is conducive to obtaining the long-range dependence of image sequences. The $D_{4}$ will restore the feature scale as the third layer by passing the upsampling operation to obtain the third feature map. The similar processing of the next layer in the decoder will be repeated, and we can obtain the fused feature $D_{i}$. The entire process can be mathematically formulated as follows:
\begin{flalign}
	D_{i}&=\begin{cases}
			\mathrm{DUM}\left ( X_{D},X_{T_{0}}  \right )  & \text{ if } i=4 \\
			\mathrm{UP}\left ( D_{i+1}  \right )   & \text{ if } i=1,2,3
		\end{cases}\\
	\mathrm{DUM}&=\mathrm{Concat}\left(\mathrm{PW}\left (X_{D},  \mathrm{gate}\left ( {X_{T}}_{0},X_{D} \right )  \right )\right )\\
	\mathrm{gate}\left ( X_{D},X_{T_{0}} \right )&= \left ( \omega_{1} \cdot X_{D}+b \right )\ast  \sigma  \left ( \omega_{2} \cdot X_{T_{0}}+c \right ).
	\label{equ_8}
\end{flalign}
\noindent where $\mathrm{PW}(\cdot)$ denotes the point-wise convolution, which is used to aggregate the features, $\mathrm{gate}$ denotes the gate unit. $\omega$ denotes the learning matrix, $b, c$ denotes the bias term, and $\sigma$ denotes the sigmod function. The $\mathrm{UP}$ operation is composed of two convolutional operations (the kernel size is 3) and bilinear interpolation with the scale factor is set to 2.

The final feature map of the last layer undergoes a convolution to restore the same size as the original input images. Then, a $1\times 1$ convolution is used to adjust the number of channels to make the final prediction.

\subsection{Loss Function}
The selection of appropriate loss functions plays a critical role in enhancing the robustness of the network. To improve the prediction accuracy of cloud image sequences, the mean squared error (MSE) loss is adopted based on the community-related works~\cite{MSTANet, STANet} to evaluate the global correlation between the ground truth (GT) $ y_{i}\in\mathbb{R} ^{T\times C\times H\times W}$ and predicted results $ y\in\mathbb{R} ^{T\times C\times H\times W}$. The formulation of the MSE loss $L_{M}$ is as follows:
\begin{flalign}
	L_{M}=\frac{1}{N} \sum_{i=1}^{N}\left(\widehat{y_{i}} -y_{i}\right)^{2}.
	\label{equ_10}
\end{flalign}
\noindent where $ N$ denotes the total number of samples.

However, the MSE loss function overlooks local structural features, which can lead to significant deviations and semantic information loss in cloud image sequence extrapolation tasks. To address this limitation, we introduce the multi-scale structural similarity (MS-SSIM) loss function to preserve edge details and structural information. The MS-SSIM is an enhanced version of the Structural Similarity Index (SSIM), incorporating structural similarity optimization across varied resolution levels. By improving robustness to scale variations in the target, MS-SSIM is particularly suitable for cloud image sequence extrapolation tasks characterized by scale-varying cloud formations.

Firstly, an $S$-level Gaussian pyramid downsampling is performed on the images $y_{i}$ and $\widehat{y_{i}}$, generating multi-scale image pairs $\left \{ y_{j}, \widehat{y_{j}} \right \} _{j=1}^{L}$ (empirically set as $L$ = 5). Three SSIM components, including luminance $l_{j}$, contrast $c_{j}$, and structure $s_{j}$ as follows:
\begin{flalign}
	l_{j} (y,\hat{y}) &=\frac{2 \mu_{y} \mu_{\hat{y}}+C_{1}}{\mu_{y}^{2}+\mu_{\hat{y}}^{2}+C_{1}} \\
	c_{j}(y, \hat{y}) &=\frac{2 \sigma_{y} \sigma_{\hat{y}}+C_{2}}{\sigma_{y}^{2}+\sigma_{\hat{y}}^{2}+C_{2}} \\
	s_{j}(y, \hat{y})&=\frac{\sigma_{y \hat{y}}+C_{3}}{\sigma_{y} \sigma_{\hat{y}}+C_{3}}.
	\label{equ_11}
\end{flalign}
\noindent where $\mu$, $\sigma$, and $\sigma_{y \hat{y}}$ denote mean, standard deviation, and covariance, respectively. $C_{1}$, $C_{2}$, and $C_{3}$ are all constants. Then, the MS-SSIM value is derived by weighted aggregation of the SSIM components across all scales:
\begin{flalign}
	\operatorname{MS-SSIM}=\left[l_{j}^{\alpha} \cdot \prod_{j=1}^{L} c_{j}^{\beta} s_{j}^{\gamma}\right].
	\label{equ_12}
\end{flalign}
\noindent where $\alpha$, $\beta$, and $\gamma$ are empirically determined weighting exponents ($\alpha$=1, $\gamma$=$\beta$=0.0448 by convention). The MS-SSIM loss $L_{MS}$ is defined as: $L_{MS}$=1-MS-SSIM.

Additionally, to emphasize the weight of the first frames in the predicted sequence, the cross-entropy (CE) loss is augmented with a weighting factor $ \tau$ (empirically set to 0.9 in this work), formulated as:
\begin{flalign}
	L_{C}=\sum_{i=1}^{T}\tau^{i}{L_{CE}}^{t+i}.
	\label{equ_13}
\end{flalign}
\noindent where $t+i$ denote the future timestep, $ L_{CE}$ denote the CE loss. Finally, the loss of our USF-Net can be formulated as: 
\begin{equation}
	L=\lambda_{1}L_{M}+\lambda_{2}L_{MS}+\lambda_{3}L_{C}.\label{equ_14}
\end{equation}
\noindent where $ \lambda_{1}=0.7$, $ \lambda_{2}=0.2$, and $ \lambda_{3}=0.1$.
\section{Experimental Results and Discussions}\label{sec:exp}
This section details the comprehensive experimental validation of USF-Net. We describe the newly created ASI-CIS dataset, the evaluation metrics, and implementation details. We then present a rigorous comparative analysis against SOTA methods, followed by extensive ablation studies to verify the contribution of each novel component.
\subsection{Dataset} \label{sec:data}
The scarcity of high-quality, large-scale public datasets remains a primary bottleneck in ground-based cloud extrapolation research. The predictive capability of deep learning models is intrinsically linked to the spatiotemporal fidelity of their training data. Existing benchmarks, such as the TSISD dataset~\cite{STANet}, utilize a constrained spatial resolution of $224 \times 224$ pixels and frequently exhibit visual occlusions arising from sensor hardware. These limitations introduce aliasing artifacts and impede the precise modeling of complex cloud dynamics.
\begin{table}[!t]
	\centering
	\caption{ASI-CIS Dataset Summary.}
	{\small
		\begin{tabular}{ccc}
			\toprule
			Weather & Size & Number / Sequences   \\
			\midrule
			Sunny & 512 $ \times $ 512 & 28,420 / 1421  \\
			Cloudy/Rainy & 512 $ \times $ 512 & 11,580 / 579  \\
			\bottomrule
	\end{tabular}}
	\label{Table0}
\end{table}

To address these challenges and provide a robust benchmark for the remote sensing community, we introduce the ASI-Cloud Image Sequence (ASI-CIS) dataset. The ASI-CIS dataset was acquired using an All-Sky Imager (ASI-DC-TK02) stationed at the meteorological observation facility in Xiqing District, Tianjin, China (geographic coordinates: $117.03^{\circ}$E, $39.10^{\circ}$N). The acquisition device features a fish-eye lens providing a hemispherical field of view, housed within a weatherproof enclosure to ensure operational continuity. The dataset consists of high-resolution images ($512 \times 512$ pixels) captured at fixed 30-second intervals. This temporal granularity is critical for capturing rapid cloud deformations and velocity changes characteristic of the lower troposphere. The data collection campaign spanned multiple seasons with daily acquisition windows from 08:00 to 17:00 local time, ensuring coverage across a diverse spectrum of solar angles and illumination conditions.
\begin{figure}[!t]
	\centering{\includegraphics[width=3.3in]{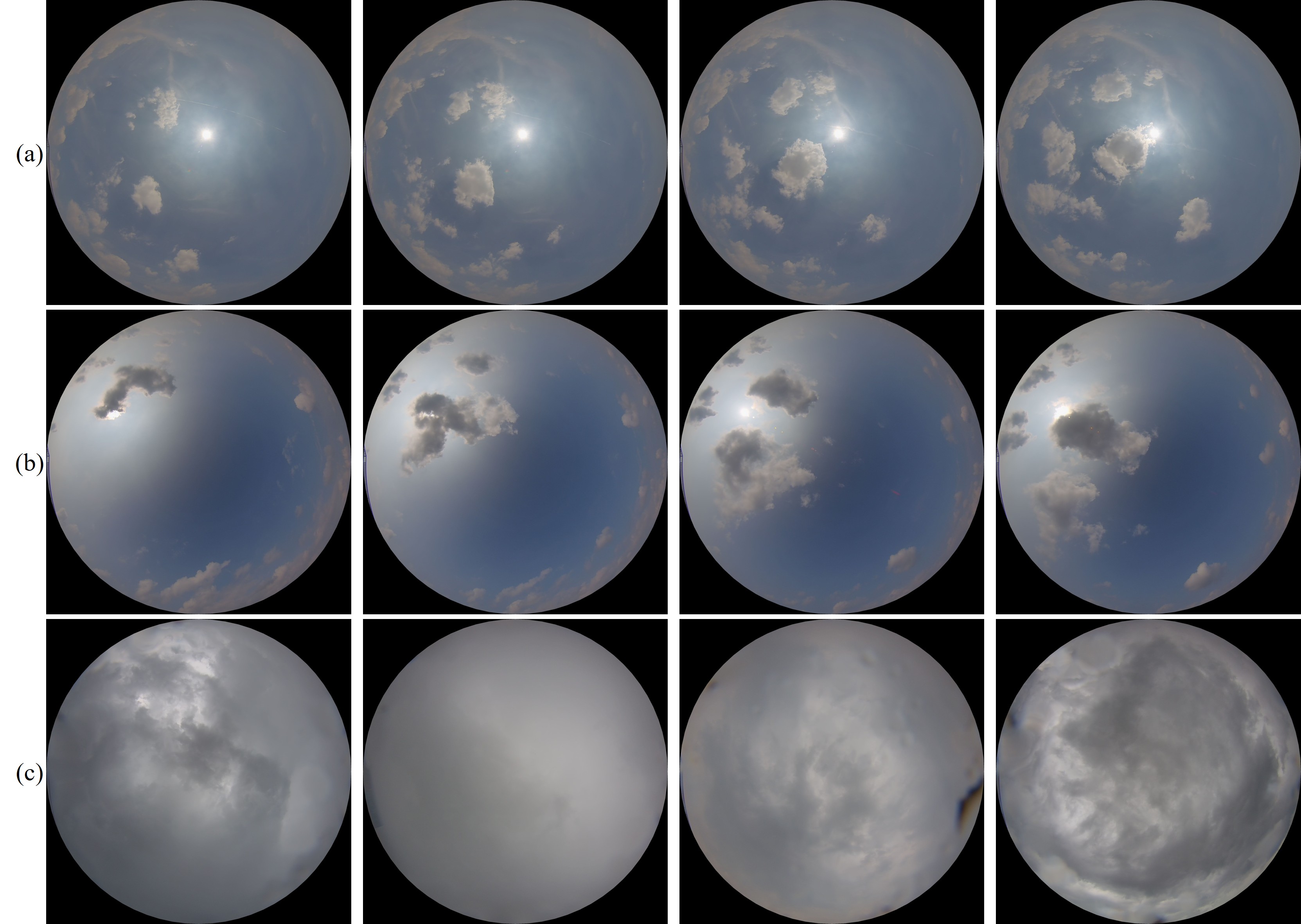}}
	\caption{(a) and (b) display valid acquisition samples under sunny and cloudy/rainy conditions, respectively; (c) displays samples rendered unsuitable for sequence extrapolation tasks due to complex cloud configurations encountered during adverse meteorological conditions such as precipitation events.}
	\label{fig_4.0}
\end{figure}

The dataset comprises a total of 2,100 sequences, partitioned into 1,400 sequences for training and 700 sequences for testing. To prevent temporal data leakage, where the model inadvertently learns from future frames closely correlated with the test set, the training and testing partitions were acquired from temporally distinct periods. Validation was conducted using a rigorous five-fold cross-validation protocol. Table \ref{Table0} summarizes the ASI-CIS dataset according to various weather conditions and quantities. Specifically, the dataset contains 1,421 sunny sequences compared to 579 cloudy/rainy sequences. While this imbalance presents a challenge for class-agnostic learning, it faithfully reflects the real-world operational environment of photovoltaic power plants in this climatic zone. Furthermore, data acquisition during precipitation events presents unique challenges. Heavy rainfall and low illumination can degrade image signal-to-noise ratios, while complex multi-layer cloud configurations (e.g., stratocumulus-cumulus mixtures) often exceed the dynamic range of standard sensors. The ASI-CIS dataset preserves these challenging samples to rigorously test model robustness under adverse meteorological conditions. Representative samples across weather conditions are shown in Fig. \ref{fig_4.0}.
\subsection{Evaluation Metrics} \label{sec:eva}
Ground-based remote sensing cloud image sequence extrapolation is fundamentally a spatiotemporal predictive learning task. To comprehensively evaluate the performance of USF-Net, we employ three standard metrics: Mean Squared Error (MSE), Structural Similarity Index (SSIM), and Peak Signal-to-Noise Ratio (PSNR).

The MSE indicates the average pixel-wise discrepancy between the predicted result and GT by computing the squared difference across corresponding pixels. The specific formulations are as follows:
\begin{flalign}
	\mathrm{MSE}=\frac{1}{m n} \sum_{i=1}^{m} \sum_{j=1}^{n}(I(i, j)-K(i, j))^{2}.
	\label{equ_15}
\end{flalign}
\noindent where $I$ and $K$ denote the predicted and GT images, respectively, and $m n$ denotes the spatial dimensions of the image.

SSIM assesses visual quality by comparing luminance, contrast, and structural similarity between images. Its value ranges from 0 (completely dissimilar) to 1 (identical), formulated as:
\begin{flalign}
	\operatorname{SSIM}(I, K)=\frac{\left(2 \mu_{I} \mu_{K}+c_{1}\right)\left(2 \sigma_{I K}+c_{2}\right)}{\left(\mu_{I}^{2}+\mu_{K}^{2}+c_{1}\right)\left(\sigma_{I}^{2}+\sigma_{K}^{2}+c_{2}\right)}.
	\label{equ_16}
\end{flalign}
\noindent where $\mu_{I}$, $\mu_{K}$ are the mean intensities; $\sigma_{I}$, $\sigma_{K}$ are the standard deviations; $\sigma_{I K}$ is the cross-covariance; and $c_{1}$, $c_{2}$are stabilization constants.

PSNR, derived from the logarithmic transformation of MSE, measures image distortion:
\begin{flalign}
	\operatorname{PSNR}=10 \cdot \log _{10}\left(\frac{\mathrm{MAX}_{I}^{2}}{\mathrm{MSE}}\right).
	\label{equ_17}
\end{flalign}
\noindent where $\mathrm{MAX}_{I}$ denotes the maximum pixel value. Higher PSNR values indicate superior reconstruction quality.
\subsection{Implementation Details} \label{sec:Imp}
The USF-Net framework was implemented using the PyTorch library and executed on a high-performance computing platform equipped with an Intel Xeon Gold 5318Y CPU (@ 2.10 GHz) and two NVIDIA A40 GPUs (48 GB VRAM).The network was optimized using the Stochastic Gradient Descent (SGD) algorithm with a momentum of 0.9 and a weight decay of $10^{-4}$ to prevent overfitting. The initial learning rate was set to $10^{-3}$. We employed a step-decay learning rate scheduler, where the learning rate is reduced by a factor of 10 every 10 epochs until reaching a minimum floor of $10^{-5}$. The training batch size was fixed at 4 samples per iteration.To mitigate exposure bias and enhance the model's capability to capture long-term spatiotemporal dynamics, we adopted a scheduled sampling strategy. The sampling probability $P$, which governs the substitution of ground truth frames with model-generated predictions during the training sequence, was linearly increased from 0 to 1 over the course of the training duration. The model was trained for 100 epochs on the ASI-CIS dataset, with a total convergence time of approximately 4.8 hours.

\subsection{Results and Discussions} \label{Res}
To evaluate the performance of our proposed method, we select several SOTA methods for comparison. These methods include both general spatiotemporal prediction methods (i.e., ConvLSTM~\cite{ConvLSTM}, PredRNN~\cite{PredRNN}, PredRNN++~\cite{PredRNN++}, MAU~\cite{MAU}, LMC~\cite{LMC}, and TAU~\cite{TAU}) and recent cloud image sequence extrapolation methods (i.e., CCLSTM~\cite{CCLSTM}, CloudPredRNN++~\cite{CloudPredRNN++}, STANet~\cite{STANet}, and MSTANet~\cite{MSTANet}). These DL methods represent different architectural paradigms (e.g., RNN-based, attention-based, hybrid models) and are widely recognized in the research community, allowing a comprehensive evaluation of USF-Net’s performance across multiple dimensions. All experimental results in this study are generated on our dataset by open-source codes.
\begin{figure*}[!t]
	\centering
	\includegraphics[width=5in]{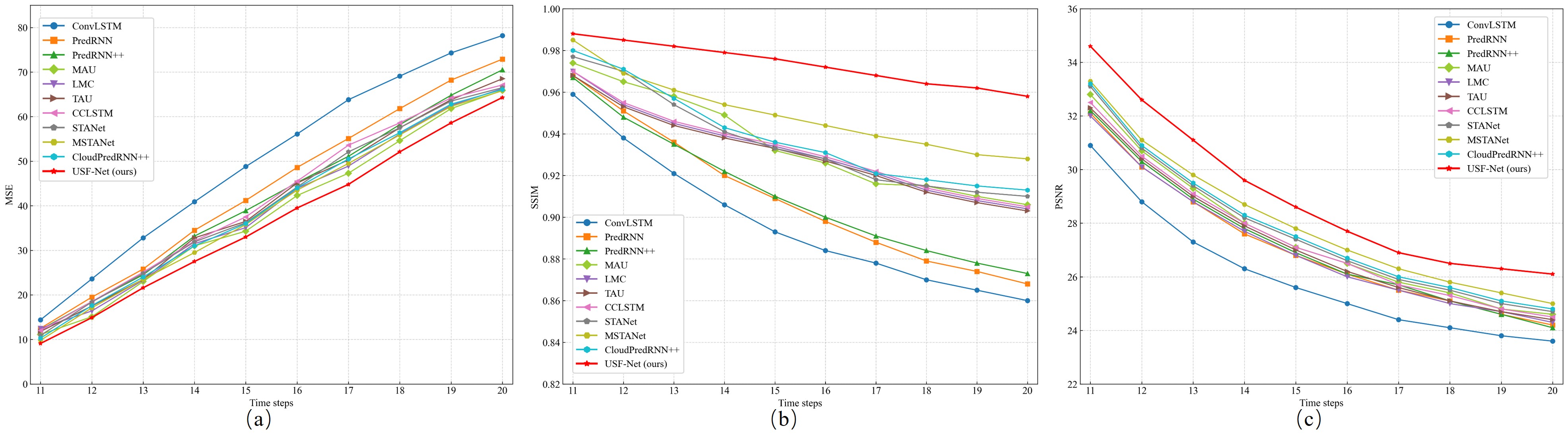}
	\caption{Quantitative timestep-by-timestep comparison between our method and other methods on three metrics (a) MSE, (b) SSIM, and (c) PSNR. From 11 to 20 are the timesteps of extrapolation in order.}
	\label{fig_4}
\end{figure*}
\subsubsection{Quantitative Comparison}
The quantitative evaluation results of our proposed USF-Net and other comparison methods on the ASI-CIS dataset are shown in Table~\ref{Table1}. Among these tables, each row is the result for each method, and each column is the metric. The highest record is marked in bold. As demonstrated in Table~\ref{Table1}, the proposed method achieves SOTA performance, attaining MSE, SSIM, and PSNR values of 37.18, 0.956, and 29.42, respectively, across all three evaluation metrics.
\begin{table}[!t]
	\centering
	\caption{Quantitative Comparison with Different Methods on the ASI-CIS Dataset. $\downarrow$ (or $\uparrow$) Indicates Lower (or Higher) is Better. The Best Results are Highlighted in Bold.}
	
	{\small
		\begin{tabular}{cccc}
			\toprule
			Method & MSE($\downarrow $) & SSIM($\uparrow $) & PSNR($\uparrow $) \\
			\midrule
			ConvLSTM (15$'$NIPS)\cite{ConvLSTM} & 50.73 & 0.887 & 25.94 \\
			PredRNN (17$'$NIPS)\cite{PredRNN} & 42.74 & 0.896 & 26.43 \\
			PredRNN++ (18$'$ICML)\cite{PredRNN++} & 41.66 & 0.911 & 26.67 \\
			MAU (21$'$NIPS)\cite{MAU} & 38.87 & 0.934 & 27.72 \\
			LMC (21$'$CVPR)\cite{LMC} & 39.67 & 0.922 & 27.13 \\
			TAU (23$'$CVPR)\cite{TAU} & 41.48 & 0.915 & 26.88 \\
			CCLSTM (21$'$RS)\cite{CCLSTM} & 39.48 & 0.929 & 27.44 \\
			STANet (23$'$TGRS)\cite{STANet} & 38.76 & 0.941 & 28.15 \\
			MSTANet (24$'$TGRS)\cite{MSTANet} & 38.11 & 0.948 & 28.66 \\
			CloudPredRNN++ (25$'$RS)\cite{CloudPredRNN++} & 38.44 & 0.945 & 28.34 \\
			USF-Net (Ours) & \textbf{37.18} & \textbf{0.956} & \textbf{29.42} \\
			\bottomrule
	\end{tabular}}
	\label{Table1}
\end{table}
To further evaluate the long-term predictive capability of our method in cloud image sequence extrapolation tasks, we present the MSE, SSIM, and PSNR of each model at every timestep. As illustrated in Fig. \ref{fig_4}, the proposed approach outperforms all baselines across metrics. The per-frame prediction curves of different models on the ASI-CIS dataset reveal distinct performance trends. Our method exhibits the weakest upward trajectory in MSE and the slowest decline in SSIM and PSNR, indicating superior stability over extended extrapolation horizons. Specifically, compared with the classic spatiotemporal sequence methods, our method introduces an SSM with a dynamic adaptive large-kernel selection mechanism, effectively addressing multi-scale variations in cloud imagery. When benchmarked against recent cloud extrapolation algorithms, our USF-Net exhibits a superior performance because the proposed UST can enhance the ability to integrate spatial and temporal features. By guiding spatial information refinement through temporal flow dynamics, the ability of our model to improve robust segmentation and capture long-term feature dependencies is enhanced. Consequently, our method achieves optimal performance even at the final timestep of extrapolation.
\subsubsection{Qualitative Comparison}
\begin{figure*}[!t]
	\centering
	\includegraphics[width=5in]{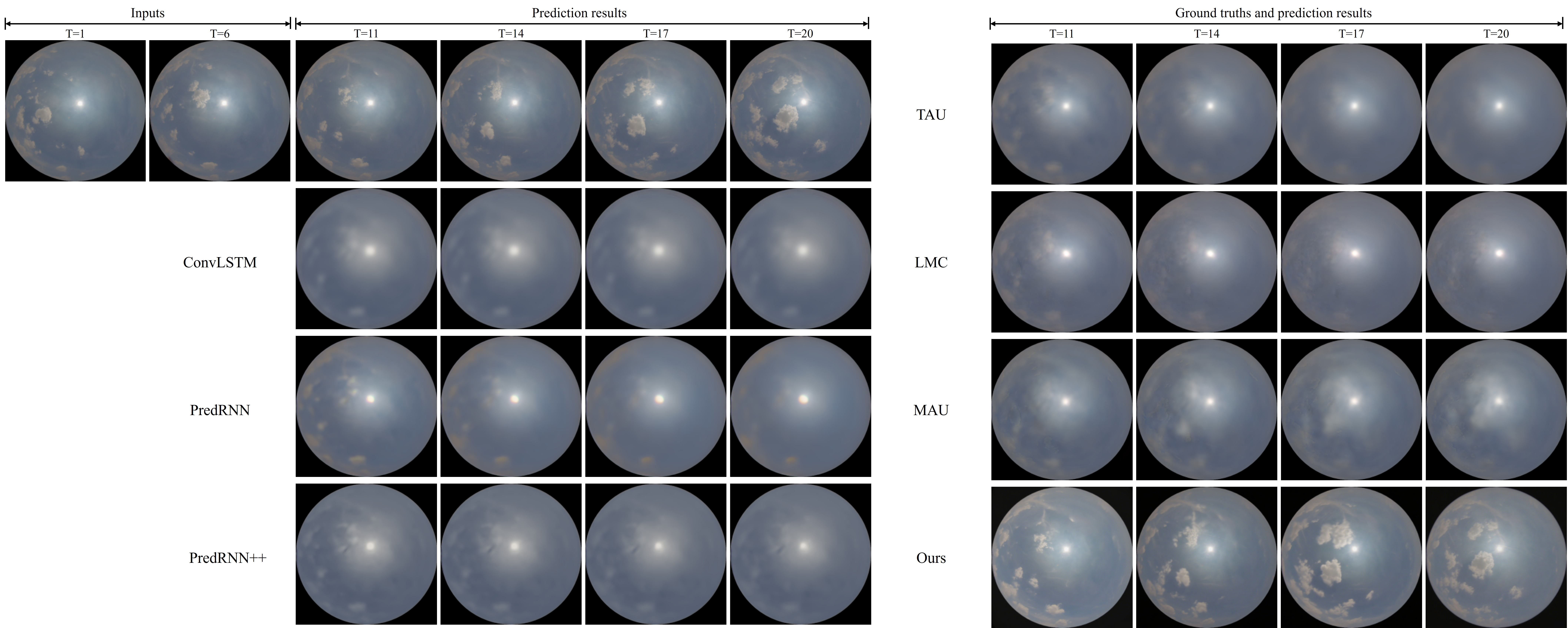}
	\caption{Comparative extrapolation performance under sunny weather conditions is presented for ConvLSTM, PredRNN, PredRNN++, TAU, LMC, MAU, and our proposed method. All experiments are conducted on the ASI-CIS dataset, predicting the next ten images given the first ten observed frames.}
	\label{fig_4.1}
\end{figure*}

\begin{figure*}[!t]
	\centering
	\includegraphics[width=5in]{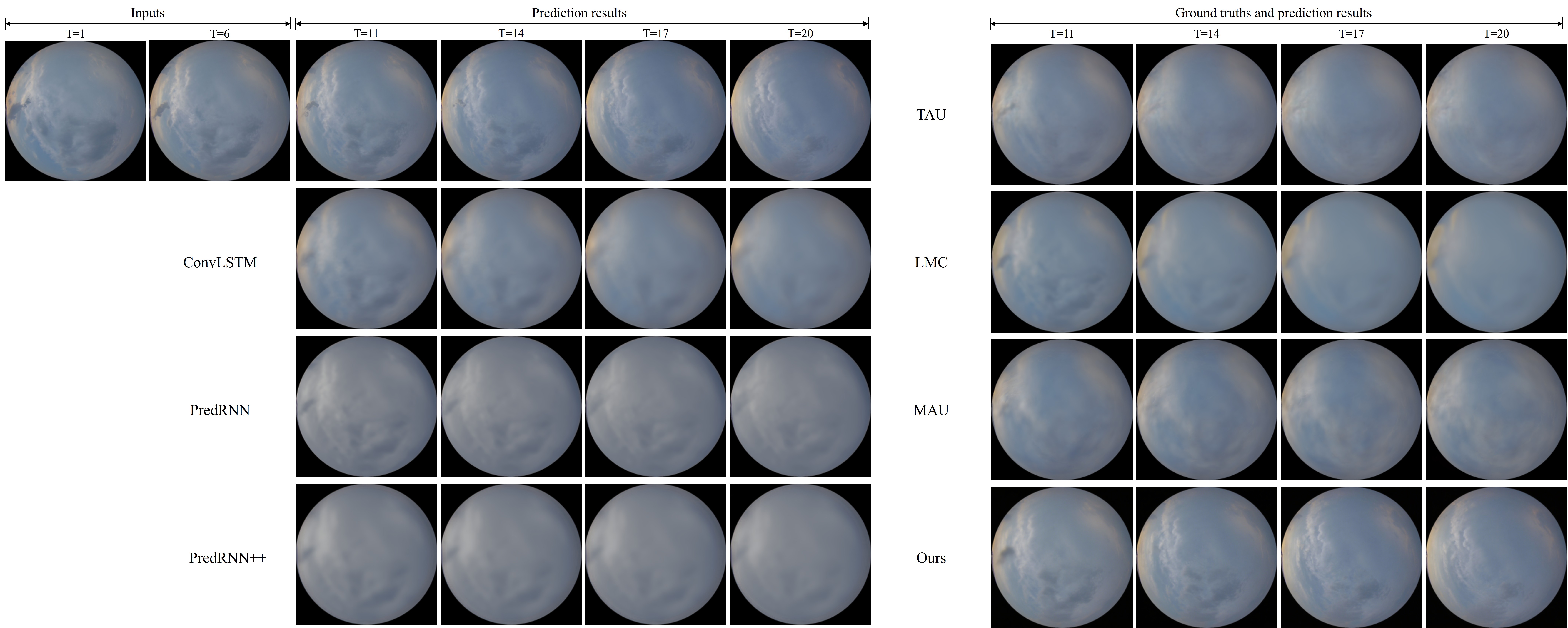}
	\caption{Comparative extrapolation performance under cloudy/rainy weather conditions is presented for ConvLSTM, PredRNN, PredRNN++, TAU, LMC, MAU, and our proposed method. All experiments are conducted on the ASI-CIS dataset, predicting the next ten images given the first ten observed frames.}
	\label{fig_4.2}
\end{figure*}

To further demonstrate the effectiveness of our method, we analyze the results of the comparison methods from a qualitative perspective. We selected some representative samples under diverse weather conditions, including sunny and cloudy/rainy scenarios, with all cloud imagery sequences exhibiting multi-scale cloud formations. Figs.~\ref{fig_4.1} -~\ref{fig_4.2} illustrate the visualization results for the representative samples. For cloud sequence extrapolation, both input and output sequences are configured with a length of 10 frames, captured at 30-second intervals. Specifically, we extracted the 1st and 6th frames (corresponding to timestamps $T$ = 1 and $T$ = 6) from each input sequence, while the 1st, 4th, 7th, and 10th output frames (corresponding to $T$ = 11, $T$ = 14, $T$ = 17, and $T$ = 20) are displayed. The first row contains the input and ground truth, and the remaining rows are the prediction results of each method.
Compared with other methods, the proposed UTS-Net exhibits a more advanced prediction performance for cloud image sequences with different scales and deformations under diverse weather conditions. As shown in Fig.~\ref{fig_4.1}, conventional temporal networks omit boundary information during sequence extrapolation in sunny scenarios. Our method preserves complete contour and boundary details, benefiting from the spatial information branch incorporated in USF-Net that captures multi-scale cloud features. Moreover, the introduced TGM significantly enhances the capability to model long-range temporal dependency. Fig.~\ref{fig_4.2} demonstrates that UTS-Net retains optimal prediction trajectories and cloud morphology even at the final extrapolation timestep. In addition, the DUM in UTS-Net effectively mitigates “ghosting effects” during sequence extrapolation as shown in Figs.~\ref{fig_4.1} -~\ref{fig_4.2}. In summary, the proposed UTS-Net exhibits robust adaptability to multi-scale cloud extrapolation tasks across varying weather patterns while achieving SOTA performance in long-term spatiotemporal modeling.
\begin{table}[!t]
	\centering
	\caption{Complexity of Different Comparative Methods on the ASI-CIS Dataset. We Report the Parameters, Flops, Inference time, and MSE. $\downarrow$ Indicates Lower is Better. The Best Results are Highlighted in Bold.}
	
	{\small
		\begin{tabular}{ccccc}
			\toprule
			Method & Params(M) & FLOPs(G) & Inference time(ms) & MSE($\downarrow $) \\
			\midrule
			ConvLSTM~\cite{ConvLSTM} & 18.0 & 215.3 & 17.3 & 50.73 \\
			PredRNN~\cite{PredRNN} & 30.5 & 382.9 & 27.9 & 42.74 \\
			PredRNN++~\cite{PredRNN++} & 48.6 & 601.1 & 28.1 & 41.66 \\
			MAU~\cite{MAU} & 19.3 & 281.1 & 16.8 & 38.72 \\
			LMC~\cite{LMC} & 20.6 & 501.1 & \textbf{14.4} & 39.67\\
			TAU~\cite{TAU} & 44.7 & 294.4 & 19.7 & 41.48\\
			CCLSTM~\cite{CCLSTM} & 55.4 & 437.1 & 51.6 & 39.48 \\
			STANet~\cite{STANet} & 26.5 & 462.9 & 16.8 & 38.76 \\
			MSTANet~\cite{MSTANet} & 24.2 & 284.3 & 16.2 & 38.11 \\
			USF-Net (Ours) & 23.8 & 266.4 & 15.8 & \textbf{37.18} \\
			\bottomrule
		\end{tabular}}
	\label{Table2}
\end{table}
\subsubsection{Complexity Comparison}
To evaluate the computation complexity of our method, we compare the parameter (Params(M)), floating-point operations (FLOPs), inference time and MSE of related methods on the ASI-CIS dataset. As shown in Table~\ref{Table2}, the proposed USF-Net achieves optimal performance with the short inference time among all evaluated methods. While our method does not exhibit advantages in parameters and FLOPs compared to classic temporal prediction methods such as ConvLSTM and MAU, its performance gains fully justify the additional computational overhead. Furthermore, our method incurs lower computational costs than attention-based cloud extrapolation methods such as STANet and MSTANet due to the integration of the TGM. As evidenced by the inference time comparison, our method achieves near-optimal efficiency (second only to LMC), which is sufficient for cloud imagery captured at 30-second intervals and aligns with the requirements of ultra-short-term PV power forecasting. Therefore, our proposed UTS-Net establishes a better accuracy-speed trade-off in ground-based remote sensing cloud image sequence extrapolation.

\subsection{Ablation Study}
To further verify the effectiveness of the different modules of our proposed UTS-Net, we also conducted a comprehensive ablation study. The proposed UTS-Net employs an encoder-decoder framework with a unified spatiotemporal module (comprising SSM-based spatial branch, TAM-based temporal branch, and TGM-based dynamic spatiotemporal module) and a decoder structure incorporating DUM. Therefore, we conduct different experiments to verify the proposed modules on the ASI-CIS dataset. First, we select UTS-Net as the baseline. Then, we incrementally remove the SSM, TAM and TGM from the baseline to verify the effectiveness of the proposed USTM. Finally, we remove the DUM from the baseline to verify its validity.
\begin{table}[!t]
	\centering
	\caption{Ablation Experimental Results on the ASI-CIS Dataset. $\downarrow$ (or $\uparrow$) Indicates Lower (or Higher) is Better. The Best Results are Highlighted in Bold.}
	{\small
		\begin{tabular}{cccccccc}
			\toprule
			Version & SSM & TAM & TGM & DUM & MSE($\downarrow $) & SSIM($\uparrow $) & Params(M)\\
			\midrule
			Baseline & \checkmark & \checkmark & \checkmark & \checkmark & \textbf{37.18} & \textbf{0.956} & 23.8\\
			1 &  & \checkmark & \checkmark & \checkmark & 39.96 & 0.918 & 23.3\\
			2 & \checkmark &  &  & \checkmark & 41.74 & 0.906 & 22.9\\
			3 & \checkmark & \checkmark &  & \checkmark & 40.24 & 0.915 & 23.1\\
			4 & \checkmark & \checkmark & \checkmark &  & 38.65 & 0.942 & 23.5\\
			5 & \checkmark & SA & \checkmark & \checkmark & 37.19 & 0.951 & 24.6\\
			\bottomrule
	\end{tabular}}
	\label{Table3}
\end{table}

\begin{figure}[!t]
	\centering{\includegraphics[width=3.3in]{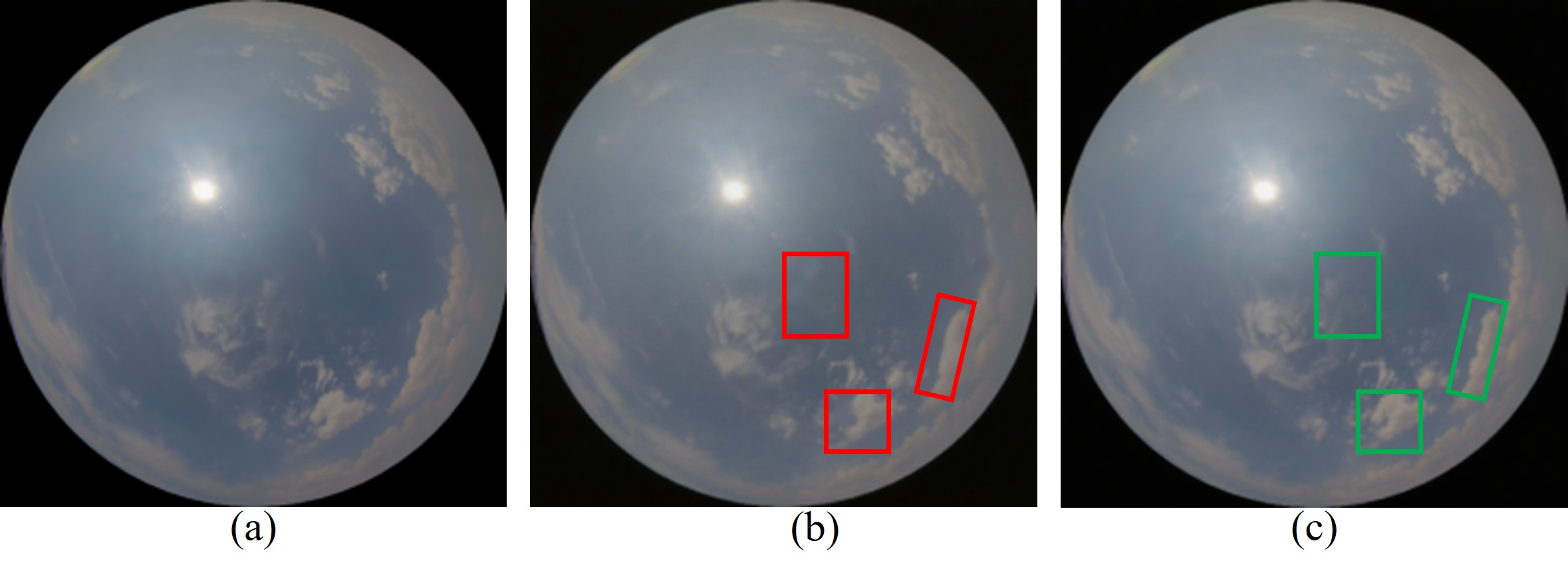}}
	\caption{To highlight the impact of the GAU, representative samples are presented: (a) ground truth data, (b) prediction from USF-Net without the DUM, and (c) prediction from USF-Net with DUM. Regions marked by red boxes indicate areas with prediction deficiencies in the absence of DUM, while green boxes demonstrate substantial improvements achieved through DUM integration.}
	\label{fig_4.3}
\end{figure}

We present a quantitative evaluation as shown in Table~\ref{Table3}. We can see that the results obtained with each module used in our UTS-Net demonstrate the effectiveness of our method. The SSM enhances the capacity of the model to extract multi-scale information about the cloud, which alleviates the pr oblem of local information loss resulting caused by scale variations in cloud imagery. By comparing the baseline and Row 1, the MSE of the model with SSM drops by 2.78$\%$ (37.18$\%$ \textit{v.s.} 39.96$\%$). It demonstrates that multi-scale contextual information plays an important role in cloud image sequence extrapolation. The introduction of the dynamic adaptive large-kernel convolution in the spatial branch improves the ability of our method to extract the topological information of clouds with variable shapes adaptively. There is a degradation of 4.56$\%$ (37.18$\%$ \textit{v.s.} 41.74$\%$) with TAM and TGM in MSE as shown in baseline and Rows 2. It demonstrates that the proposed temporal-guided spatial refinement mechanism enhances the capability of the network to capture global relationships of information between different stages and the correlations of long-range features. In addition, by comparing the baseline and Row 4, the SSIM of the method without DUM drops by 1.4$\%$ (95.6$\%$ v.s. 94.2$\%$). It demonstrates that the decoder with DUM effectively alleviates information loss between the encoder and decoder, reducing “ghosting effect” and improving extrapolation fidelity. As illustrated in Fig.~\ref{fig_4.3}, the proposed USF-Net incorporating the DUM demonstrates superior predictive performance compared to its DUM-free counterpart, demonstrating the module's efficacy in mitigating “ghosting effects” commonly encountered in the ground-based remote sensing cloud image sequence extrapolation tasks. Finally, as shown in the last row of Table~\ref{Table3}, the TAM reduces parameters by 0.8 compared to conventional self-attention (SA) mechanisms, with also marginal MSE degradation (23.8 \textit{v.s.} 24.6). This confirms that the temporal branch with TAM achieves computational efficiency while preserving long-term temporal dependency modeling.
\begin{table}[!t]
	\centering
	\caption{Ablation Experimental Results of the Number of Decomposed Large Kernels with the RF being 23.}
	{\small
		\begin{tabular}{ccccc}
			\toprule
			RF & (k,d) Sequence & Number & Inference time(ms) & MSE($\downarrow $) \\
			\midrule
			23 & (23, 1) & 1 & 16.6 & 38.21\\
			23 & (5, 1)$\longrightarrow $(7, 3) & 2 & \textbf{15.8} & \textbf{37.18}\\
			23 & (3, 1)$\longrightarrow $(5, 1)$\longrightarrow $(7, 2) & 3 & 15.4 & 37.53\\
			
			\bottomrule
	\end{tabular}}
	\label{Table4}
\end{table}
\begin{table}[!t]
	\centering
	\caption{Ablation Experimental Results with Different RFs of the Dynamic Large-kernel Selection. RF = 23 Corresponds to Our Proposed Method.}
	{\small
		\begin{tabular}{ccccc}
			\toprule
			RF & (k,d) Sequence & Inference time(ms) & Params(M) & MSE($\downarrow $) \\
			\midrule
			11 & (3, 1)$\longrightarrow $(5, 2) & 17.2 & 22.1 & 39.65\\
			21 & (3, 1)$\longrightarrow $(7, 3) & 16.1 & 23.4 & 37.64\\
			23 & (5, 1)$\longrightarrow $(7, 3) & \textbf{15.8} & \textbf{23.8} & \textbf{37.18}\\
			29 & (5, 1)$\longrightarrow $(7, 4) & 15.6 & 24.4 & 37.47\\
			39 & (7, 1)$\longrightarrow $(9, 4) & 16.3 & 25.6 & 38.14\\
			\bottomrule
	\end{tabular}}
	\label{Table5}
\end{table}
Moreover, to evaluate the impact of dynamic large-kernel selection on cloud image sequence extrapolation performance, the ablation study is conducted on the selection of the multi-scale large-kernel in the spatial branch. When the RF is fixed at 23, we conduct an experiment on the number of large kernel decompositions. The experimental results, as shown in Table~\ref{Table4}, achieve a good trade-off between speed and accuracy by decomposing the large kernel into two depth-wise kernels, resulting in excellent performance in both inference time and MSE. In addition, we configured the RF as 11, 21, 23, 29, and 39, where RF = 23 corresponds to our proposed method. As shown in Table~\ref{Table5}, decomposing large kernels into two depth-wise components effectively captures multi-scale cloud motion patterns, significantly improving prediction accuracy for cloud image sequences. However, excessively small or large RFs can hinder the performance of the USF-Net. The performance degrades when the RF exceeds 23 due to excessive detail loss when decomposed kernels encounter smaller-scale cloud structures. The experimental results demonstrate that our selected large kernel decomposition strategy achieves an optimal balance between prediction performance and computational efficiency.

\section{Conclusion}\label{sec:conclusion}
Precise and efficient extrapolation of ground-based cloud image sequences constitutes a pivotal enabling technology for mitigating the intermittency inherent in photovoltaic power integration. This study circumvents the limitations of prevailing deep learning methodologies by introducing USF-Net, a novel framework that orchestrates a spatiotemporal architecture to synthesize spatial features with temporal dynamics. The primary technical contributions are realized through three specialized components. First, a TGM was developed to explicitly modulate spatial feature learning using temporal flow, thereby ensuring coherent spatiotemporal representations. Second, the USTM was engineered, comprising a SSM for dynamic multi-scale context capture and a TAM that resolves long-range dependencies with linear computational complexity. Third, a DUM integrated into the decoder leverages initial temporal states to mitigate the ghosting effect, preserving high-fidelity motion signatures. Extensive empirical validation on the newly curated high-resolution ASI-CIS dataset demonstrates that USF-Net surpasses current SOTA benchmarks. Concurrently, comprehensive ablation studies corroborate the individual efficacy of the proposed TGM, USTM, and DUM components. By achieving a superior equilibrium between prediction accuracy and computational efficiency, USF-Net establishes a new benchmark for the domain. Future research will explore the extensibility of USF-Net to broader photovoltaic power forecasting tasks and the further optimization of real-time inference capabilities.

\bmhead{Acknowledgements}
The research was supported by the National Natural Science Foundation of China under Grant No. 62206085, supported by the Innovation Capacity Improvement Plan Project of Hebei Province under 22567603H, and supported by the Interdisciplinary Postgraduate Training Program of Hebei University of Technology under HEBUT-Y-XKJC-2022101.





\bibliography{sn-bibliography}

@article{Intro1,
  author={Shi, Jie and Lee, Wei-Jen and Liu, Yongqian and Yang, Yongping and Wang, Peng},
  journal={{IEEE} Trans. Ind. Appl.	}, 
  title={Forecasting Power Output of Photovoltaic Systems Based on Weather Classification and Support Vector Machines}, 
  year={2012},
  volume={48},
  number={3},
  pages={1064-1069},
  doi={10.1109/TIA.2012.2190816}
  }

@article{Intro2,
title = {3D cloud detection and tracking system for solar forecast using multiple sky imagers},
journal = {Sol. Energy},
volume = {118},
pages = {496-519},
year = {2015},
author = {Zhenzhou Peng and Dantong Yu and Dong Huang and John Heiser and Shinjae Yoo and Paul Kalb},
doi = {https://doi.org/10.1016/j.solener.2015.05.037},
}

@article{Intro3,
title = {ConvODE-Mixer: A multimodal deep learning model for ultra-short-term PV power forecasting},
journal = {Sol. Energy},
volume = {300},
pages = {113777},
year = {2025},
issn = {0038-092X},
doi = {https://doi.org/10.1016/j.solener.2025.113777},
author = {Binbin Yong and Yanxiang Zhang and Jun Shen and Aiai Ren and Xu Zhou and Qingguo Zhou}}

@article{Intro4,
  author       = {Bo Zhong and
                  Wuhan Chen and
                  Shanlong Wu and
                  Longfei Hu and
                  Xiaobo Luo and
                  Qinhuo Liu},
  title        = {A Cloud Detection Method Based on Relationship Between Objects of
                  Cloud and Cloud-Shadow for Chinese Moderate to High Resolution Satellite
                  Imagery},
  journal      = {{IEEE} J. Sel. Top. Appl. Earth Obs. Remote. Sens.},
  volume       = {10},
  number       = {11},
  pages        = {4898--4908},
  year         = {2017},
  doi          = {10.1109/JSTARS.2017.2734912}

}

@article{Intro5,
title = {Cloud detection method based on clear sky background under multiple weather conditions},
journal = {Sol. Energy},
volume = {255},
pages = {1-11},
year = {2023},
issn = {0038-092X},
doi = {https://doi.org/10.1016/j.solener.2023.03.026},
author = {Jifeng Song and Zixuan Yan and Yisen Niu and Lianglin Zou and Xilong Lin}
}

@article{Intro6,
title = {CloudSwinNet: A hybrid CNN-transformer framework for ground-based cloud images fine-grained segmentation},
journal = {Energy},
volume = {309},
pages = {133128},
year = {2024},
issn = {0360-5442},
doi = {https://doi.org/10.1016/j.energy.2024.133128},
author = {Chaojun Shi and Zibo Su and Ke Zhang and Xiongbin Xie and Xiaoyun Zhang}
}

@article{25TIM,
  author       = {Bing Nie and
                  Zhiying Lu and
                  Jun Han and
                  Wenpeng Chen and
                  Chao Cai and
                  Wenjie Pan},
  title        = {Investigation on Ground-Based Cloud Image Classification and Its Application
                  in Photovoltaic Power Forecasting},
  journal      = {{IEEE} Trans. Instrum. Meas.},
  volume       = {74},
  pages        = {1--11},
  year         = {2025},
  doi          = {10.1109/TIM.2025.3529074}
}

@article{Intro7,
title = {Research on ultra-short-term photovoltaic power forecasting using multimodal data and ensemble learning},
journal = {Energy},
volume = {330},
pages = {136831},
year = {2025},
issn = {0360-5442},
doi = {https://doi.org/10.1016/j.energy.2025.136831},
author = {Yifeng Ma and Wenzheng Yu and Junyu Zhu and Zhiyuan You and Aiqing Jia}}

@article{Intro8,
title = {A multi-modal deep clustering method for day-ahead solar irradiance forecasting using ground-based cloud imagery and time series data},
journal = {Energy},
volume = {321},
pages = {135285},
year = {2025},
issn = {0360-5442},
doi = {https://doi.org/10.1016/j.energy.2025.135285},
author = {Weijing Dou and Kai Wang and Shuo Shan and Mingyu Chen and Kanjian Zhang and Haikun Wei and Victor Sreeram}}

@article{Intro9,
title = {Convolutional neural networks for intra-hour solar forecasting based on sky image sequences},
journal = {Appl. Energy},
volume = {310},
pages = {118438},
year = {2022},
issn = {0306-2619},
doi = {https://doi.org/10.1016/j.apenergy.2021.118438},
author = {Cong Feng and Jie Zhang and Wenqi Zhang and Bri-Mathias Hodge}}

@incollection{Intro10,
  author       = {H. Guo and
                  Anand Rangarajan and
                  Shantanu H. Joshi},
  booktitle    = {Handbook of Mathematical Models in Computer Vision},
  pages        = {205--219},
  year         = {2006},
  doi          = {10.1007/0-387-28831-7\_13},
  bibsource    = {dblp computer science bibliography, https://dblp.org}
}

@article{Intro11,
title = {A hybrid approach to estimate the complex motions of clouds in sky images},
journal = {Sol. Energy},
volume = {138},
pages = {10-25},
year = {2016},
issn = {0038-092X},
doi = {https://doi.org/10.1016/j.solener.2016.09.002},
author = {Zhenzhou Peng and Dantong Yu and Dong Huang and John Heiser and Paul Kalb}
}

@article{Intro12,
  author       = {Michael H{\"{u}}sken and
                  Peter Stagge},
  title        = {Recurrent neural networks for time series classification},
  journal      = {Neurocomputing},
  volume       = {50},
  pages        = {223--235},
  year         = {2003},
  doi          = {10.1016/S0925-2312(01)00706-8},
  bibsource    = {dblp computer science bibliography, https://dblp.org}
}

@article{Intro14,
  author={Hochreiter, Sepp and Schmidhuber, Jürgen},
  journal={Neural Comput.}, 
  title={Long Short-Term Memory}, 
  year={1997},
  volume={9},
  number={8},
  pages={1735-1780},
  doi={10.1162/neco.1997.9.8.1735}
}

@inproceedings{PredRNN,
  author       = {Yunbo Wang and
                  Mingsheng Long and
                  Jianmin Wang and
                  Zhifeng Gao and
                  Philip S. Yu},
  title        = {PredRNN: Recurrent Neural Networks for Predictive Learning using Spatiotemporal
                  LSTMs},
  booktitle    = {Proc. Adv. neural inf. proces. syst. (NIPS)},
  address	   = {Long Beach, CA, USA},
  month     = {Dec.},
  pages        = {879--888},
  year         = {2017}
}

@InProceedings{PredRNN++,
  title = 	 {{P}red{RNN}++: Towards A Resolution of the Deep-in-Time Dilemma in Spatiotemporal Predictive Learning},
  author =       {Wang, Yunbo and Gao, Zhifeng and Long, Mingsheng and Wang, Jianmin and Yu, Philip S},
  booktitle = 	 {Int. Conf. Mach. Learn. (ICML)},
  address = {Stockholm, Sweden},
  pages = 	 {5123--5132},
  year = 	 {2018},
  volume = 	 {80},
  month = 	 {Jul.}
}

@inproceedings{LMC,
  author       = {Wei Yu and
                  Yichao Lu and
                  Steve Easterbrook and
                  Sanja Fidler},
  title        = {Efficient and Information-Preserving Future Frame Prediction and Beyond},
  booktitle    = {Int. Conf. Learn. Represent. (ICLR)},
  address = {Addis Ababa, Ethiopia},
  year = 	 {2020},
  month = 	 {Apr.}
}

@inproceedings{ConvLSTM,
  author       = {Xingjian Shi and
                  Zhourong Chen and
                  Hao Wang and
                  Dit{-}Yan Yeung and
                  Wai{-}Kin Wong and
                  Wang{-}chun Woo},
  title        = {Convolutional {LSTM} Network: {A} Machine Learning Approach for Precipitation
                  Nowcasting},
  booktitle    = {Proc. Adv. neural inf. proces. syst. (NIPS)},
  address = {Montreal, Quebec, Canada}, 
  month = {Dec.},
  pages        = {802--810},
  year         = {2015}
}

@article{Intro16,
title = {On the use of sky images for intra-hour solar forecasting benchmarking: Comparison of indirect and direct approaches},
journal = {Sol. Energy},
volume = {276},
pages = {112649},
year = {2024},
issn = {0038-092X},
doi = {https://doi.org/10.1016/j.solener.2024.112649},
author = {Guoping Ruan and Xiaoyang Chen and Eng Gee Lim and Lurui Fang and Qi Su and Lin Jiang and Yang Du}}

@article{Intro17,
title = {A radiant shift: Attention-embedded CNNs for accurate solar irradiance forecasting and prediction from sky images},
journal = {Renewable Energy},
volume = {234},
pages = {121133},
year = {2024},
issn = {0960-1481},
doi = {https://doi.org/10.1016/j.renene.2024.121133},
author = {Anto Leoba Jonathan and Dongsheng Cai and Chiagoziem C. Ukwuoma and Nkou Joseph Junior Nkou and Qi Huang and Olusola Bamisile}}

@article{MSTANet,
  author       = {Feng Zhang and
                  Yingying Cheng and
                  Qiang Hua and
                  Chunru Dong and
                  Yong Zhang and
                  Tingdong Wu},
  title        = {A Multiscale Spatiotemporal Attention Network for Ground-Based Remote
                  Sensing Cloud Image Sequence Prediction},
  journal      = {{IEEE} Trans. Geosci. Remote. Sens.},
  volume       = {62},
  pages        = {1--13},
  year         = {2024},
  doi          = {10.1109/TGRS.2024.3485581}
}

@inproceedings{MAU,
  author       = {Zheng Chang and
                  Xinfeng Zhang and
                  Shanshe Wang and
                  Siwei Ma and
                  Yan Ye and
                  Xiang Xinguang and
                  Wen Gao},
  title        = {{MAU:} {A} Motion-Aware Unit for Video Prediction and Beyond},
  booktitle    = {Proc. Adv. neural inf. proces. syst. (NIPS)}, 
  address = {Virtual, Online},
  month = {Dec.},
  pages        = {26950--26962},
  year         = {2021}
}

@InProceedings{LSK,
    author    = {Li, Yuxuan and Hou, Qibin and Zheng, Zhaohui and Cheng, Ming-Ming and Yang, Jian and Li, Xiang},
    title     = {Large Selective Kernel Network for Remote Sensing Object Detection},
    booktitle = {Proc. IEEE. Int. Conf. Comput. Vision. (ICCV)},
	address   = {Paris, France},
    month     = {Oct.},
    year      = {2023},
    pages     = {16794-16805},
    doi       = {10.1109/ICCV51070.2023.01540}
}

@article{STANet,
  author       = {Zhiying Lu and
                  Zhiyi Zhou and
                  Xin Li and
                  Jianfeng Zhang},
  title        = {STANet: {A} Novel Predictive Neural Network for Ground-Based Remote
                  Sensing Cloud Image Sequence Extrapolation},
  journal      = {{IEEE} Trans. Geosci. Remote. Sens.},
  volume       = {61},
  pages        = {1--11},
  year         = {2023},
  doi          = {10.1109/TGRS.2023.3268503}
}

@Article{CloudPredRNN++,
AUTHOR = {Li, Sheng and Wang, Min and Shi, Minghang and Wang, Jiafeng and Cao, Ran},
TITLE = {Leveraging Deep Spatiotemporal Sequence Prediction Network with Self-Attention for Ground-Based Cloud Dynamics Forecasting},
JOURNAL = {Remote Sens.},
VOLUME = {17},
YEAR = {2025},
NUMBER = {1},
ARTICLE-NUMBER = {18},
ISSN = {2072-4292},
DOI = {10.3390/rs17010018}
}

@INPROCEEDINGS{Re1,
  author={El Jaouhari, Zakaria and Zaz, Youssef and Masmoudi, Lhoussain},
  booktitle={Proc. IEEE Int. Renew. Sustain. Energy Conf. (IRSEC)}, 
  title={Cloud tracking from whole-sky ground-based images},
  address	   = {Marrakech, Morocco},
  month		   = {Dec.},
  year={2015},
  pages={1-5}
  
}

@Article{Re2,
AUTHOR = {Du, Juan and Min, Qilong and Zhang, Penglin and Guo, Jinhui and Yang, Jun and Yin, Bangsheng},
TITLE = {Short-Term Solar Irradiance Forecasts Using Sky Images and Radiative Transfer Model},
JOURNAL = {Energies},
VOLUME = {11},
YEAR = {2018},
NUMBER = {5},
ARTICLE-NUMBER = {1107},
ISSN = {1996-1073},
DOI = {10.3390/en11051107}
}

@inproceedings{Re3,
Author = {Chang, Ming-Ching and Yao, Yi and Li, Guan and Tong, Yan and Tu, Peter},
Title = {CLOUD TRACKING FOR SOLAR IRRADIANCE PREDICTION},
Booktitle = {Proc. Int. Conf. Image Process. (ICIP)},
address	   = {Beijing, China},
month		   = {Sep.},
Year = {2017},
Pages = {4387-4391},
doi   = {10.1109/ICIP.2017.8297111}

}

@article{Re4,
title = {Image phase shift invariance based cloud motion displacement vector calculation method for ultra-short-term solar PV power forecasting},
journal = {Energy Convers. Manage.},
volume = {157},
pages = {123-135},
year = {2018},
issn = {0196-8904},
doi = {https://doi.org/10.1016/j.enconman.2017.11.080},
author = {Fei Wang and Zhao Zhen and Chun Liu and Zengqiang Mi and Bri-Mathias Hodge and Miadreza Shafie-khah and João P.S. Catalão}
}

@Article{Re5,
AUTHOR = {Ye, Yuankang and Gao, Feng and Cheng, Wei and Liu, Chang and Zhang, Shaoqing},
TITLE = {MSSTNet: A Multi-Scale Spatiotemporal Prediction Neural Network for Precipitation Nowcasting},
JOURNAL = {Remote Sens.},
VOLUME = {15},
YEAR = {2023},
NUMBER = {1},
ARTICLE-NUMBER = {137},
ISSN = {2072-4292},
DOI = {10.3390/rs15010137}
}

@inproceedings{Re6,
title={Eidetic 3D {LSTM}: A Model for Video Prediction and Beyond},
author={Yunbo Wang and Lu Jiang and Ming-Hsuan Yang and Li-Jia Li and Mingsheng Long and Li Fei-Fei},
booktitle={Int. Conf. Learn. Represent. (ICLR)},
address	   = {New Orleans, LA, USA},
month		   = {May.},
Year = {2019},
Pages = {41}
}

@InProceedings{Re7,
    author    = {Wu, Haixu and Yao, Zhiyu and Wang, Jianmin and Long, Mingsheng},
    title     = {MotionRNN: A Flexible Model for Video Prediction With Spacetime-Varying Motions},
    booktitle = {Proc. IEEE Conf. Comput. Vis. Pattern Recognit. (CVPR)},
    address	   = {Virtual, Online, USA},
	month     = {Jun.},
    year      = {2021},
    pages     = {15435-15444},
	doi          = {10.1109/CVPR46437.2021.01518}
}

@ARTICLE{Re8,
  author       = {Cheng Tan and
                  Zhangyang Gao and
                  Siyuan Li and
                  Stan Z. Li},
  title        = {SimVPv2: Towards Simple Yet Powerful Spatiotemporal Predictive Learning},
  journal      = {{IEEE} Trans. Multim.},
  volume       = {27},
  pages        = {5170--5184},
  year         = {2025},
  doi          = {10.1109/TMM.2025.3543051}
}

@InProceedings{TAU,
    author    = {Tan, Cheng and Gao, Zhangyang and Wu, Lirong and Xu, Yongjie and Xia, Jun and Li, Siyuan and Li, Stan Z.},
    title     = {Temporal Attention Unit: Towards Efficient Spatiotemporal Predictive Learning},
    booktitle = {Proc. IEEE Conf. Comput. Vis. Pattern Recognit. (CVPR)},
    address	   = {Vancouver, BC, Canada},
	month     = {Jun.},
    year      = {2023},
    pages     = {18770-18782},
	doi          = {10.1109/CVPR52729.2023.01800}
}

@Article{CCLSTM,
AUTHOR = {Lu, Zhiying and Wang, Zehan and Li, Xin and Zhang, Jianfeng},
TITLE = {A Method of Ground-Based Cloud Motion Predict: CCLSTM + SR-Net},
JOURNAL = {Remote Sens.},
VOLUME = {13},
YEAR = {2021},
NUMBER = {19},
ARTICLE-NUMBER = {3876},
ISSN = {2072-4292},
DOI = {10.3390/rs13193876}
}

\end{document}